\PassOptionsToPackage{frozencache}{minted}
\documentclass[dvipsnames]{article}

\usepackage[final, nonatbib]{neurips_2024}

\usepackage[utf8]{inputenc} 
\usepackage[T1]{fontenc}    
\usepackage{hyperref}       
\usepackage{url}            
\usepackage{booktabs}       
\usepackage{amsfonts}       
\usepackage{nicefrac}       
\usepackage{microtype}      
\usepackage{xcolor}         
\usepackage[numbers]{natbib}

\usepackage{tabularx}
\usepackage{tabularray}
\usepackage{listings}
\usepackage[inline]{enumitem}
\usepackage{caption}
\usepackage{subcaption}
\usepackage{graphicx}
\usepackage{glossaries}
\usepackage{adjustbox}
\usepackage{pgfplots}
\usepackage{mleftright}
\usepackage{derivative}
\usepackage{pifont}
\usepackage{titlesec}
\usepackage{wrapfig}
\usepackage{fvextra}
\usepackage{algorithm,algorithmicx,algpseudocode}
\usepackage{listings}
\usepackage[many]{tcolorbox}
\tcbuselibrary{minted,skins}

\UseTblrLibrary{booktabs}
\definecolor{myBlue}{HTML}{0F1A5F}
\definecolor{myRed}{HTML}{721010}
\hypersetup{
  colorlinks,
  linkcolor={myRed},
  citecolor={myBlue},
  urlcolor={blue!80!black}
  }
\glsdisablehyper
\hypersetup{breaklinks=true}
\usepgfplotslibrary{groupplots}
\usepgfplotslibrary{colormaps}
\usepgfplotslibrary{fillbetween}
\usetikzlibrary{plotmarks}

\pgfplotsset{compat=1.14}	 %
\pgfplotsset{compat/show suggested version=false}
\pgfplotsset{every mark/.append style={solid}}
\mleftright

\newcommand{\cmark}{\ding{51}}%
\newcommand{\xmark}{\ding{55}}%

\titlespacing*{\paragraph}{0pt}{0.35\baselineskip}{1em}

\definecolor{contessa}{HTML}{BF616A}
\definecolor{LG}{gray}{0.95}
\DeclareTCBListing[blend into=figures,]{mintedbox}{O{}mO{}}{%
enhanced,
  listing only,
  breakable,
  minted language=#2,
  minted options={%
    frame=lines,
    framesep=2mm,
    baselinestretch=1.2,
    linenos,
    numbersep=5pt,
    gobble=0,
    tabsize=4,
    breaklines=true,
    mathescape,
    #1},
  top=0pt,
  bottom=0pt,
  left=0pt,
  right=0pt,
  arc=0pt,
  colframe=white,
  colback=LG,
  coltitle=black,
  colbacktitle=white,
  #3}

\usepackage{amsmath}
\usepackage{amssymb}
\usepackage{mathtools}
\usepackage{amsthm}

\usepackage[capitalize,noabbrev]{cleveref}

\theoremstyle{plain}

\theoremstyle{definition}

\theoremstyle{remark}

\definecolor{C0}{rgb}{0.121569, 0.466667, 0.705882}
\definecolor{C1}{rgb}{1.000000, 0.498039, 0.054902}
\definecolor{C2}{rgb}{0.172549, 0.627451, 0.172549}
\definecolor{C3}{rgb}{0.839216, 0.152941, 0.156863}
\definecolor{C4}{rgb}{0.580392, 0.403922, 0.741176}
\definecolor{C5}{rgb}{0.549020, 0.337255, 0.294118}
\definecolor{C6}{rgb}{0.890196, 0.466667, 0.760784}
\definecolor{C7}{rgb}{0.498039, 0.498039, 0.498039}
\definecolor{C8}{rgb}{0.737255, 0.741176, 0.133333}
\definecolor{C9}{rgb}{0.090196, 0.745098, 0.811765}

\newcolumntype{Y}{>{\centering\arraybackslash}X}
\newcolumntype{C}{>{\hsize=.0\hsize\centering\arraybackslash}X}

\colorlet{LightGoldenrod}{White!40!Goldenrod}
\colorlet{LightGray}{White!90!Periwinkle}

\definecolor{codegreen}{rgb}{0,0.6,0}
  \definecolor{codegray}{rgb}{0.5,0.5,0.5}
  \definecolor{codepurple}{rgb}{0.58,0,0.82}
  \definecolor{backcolour}{rgb}{0.95,0.95,0.92}
  \lstdefinestyle{mystyle}{
    backgroundcolor=\color{backcolour},
    commentstyle=\color{codegreen},
    keywordstyle=\color{magenta},
    numberstyle=\tiny\color{codegray},
    stringstyle=\color{codepurple},
    basicstyle=\ttfamily\footnotesize,
    breakatwhitespace=false,
    breaklines=true,
    captionpos=b,
    keepspaces=true,
    numbers=left,
    numbersep=5pt,
    showspaces=false,
    showstringspaces=false,
    showtabs=false,
    tabsize=2
  }
  
  \lstnewenvironment{mylisting}{\lstset{style=mystyle}}{}

\newcommand{\method}{{LiteVAE}}

\usepackage{amsmath,amsfonts,bm, mathtools}

\newcommand{\mc}[1]{\mathcal{#1}}

\def\eqref#1{equation~\ref{#1}}

\def\1{\bm{1}}

\def\beps{\bm{\epsilon}}

\def\vx{{\bm{x}}}

\def\vz{{\bm{z}}}

\def\mI{{\bm{I}}}

\def\bmtheta{{\bm{\theta}}}

\DeclareMathAlphabet{\mathsfit}{\encodingdefault}{\sfdefault}{m}{sl}
\SetMathAlphabet{\mathsfit}{bold}{\encodingdefault}{\sfdefault}{bx}{n}

\newcommand{\pdata}{p_{\rm{data}}}

\newcommand{\norm}[1]{\left\lVert #1 \right\rVert}

\newcommand{\trp}[1]{#1^{\top}}

\newcommand{\normal}[2]{\mc{N}\prn{#1, #2}}

\DeclarePairedDelimiterX{\infdivx}[2]{(}{)}{%
  #1\delimsize\|#2%
}

\newcommand{\prn}[1]{\left( #1 \right)}

\newcommand{\zero}{\pmb{0}}

\DeclareDocumentCommand{\ex}{m o}{
   \mathbb{E}\IfValueT{#2}{_{#2}}\left[#1\right]
}

\DeclareMathOperator{\grad}{\nabla}

\DeclarePairedDelimiterX\Set[1]{\lbrace}{\rbrace}%
 {  #1 }

\def\ddefloop#1{\ifx\ddefloop#1\else\ddef{#1}\expandafter\ddefloop\fi}
\def\ddef#1{\expandafter\def\csname #1bb\endcsname{\ensuremath{\mathbb{#1}}}}
\ddefloop ABCDEFGHIJKLMNOPQRSTUVWXYZ\ddefloop
\def\ddefloop#1{\ifx\ddefloop#1\else\ddef{#1}\expandafter\ddefloop\fi}
\def\ddef#1{\expandafter\def\csname #1b\endcsname{\ensuremath{\mathbf{#1}}}}
\ddefloop ABCDEFGHIJKLMNOPQRSTUVWXYZ\ddefloop
\def\ddef#1{\expandafter\def\csname #1c\endcsname{\ensuremath{\mathcal{#1}}}}
\ddefloop ABCDEFGHIJKLMNOPQRSTUVWXYZ\ddefloop
\def\ddef#1{\expandafter\def\csname #1hat\endcsname{\ensuremath{\widehat{#1}}}}
\ddefloop ABCDEFGHIJKLMNOPQRSTUVWXYZ\ddefloop
\def\ddef#1{\expandafter\def\csname hc#1\endcsname{\ensuremath{\widehat{\mathcal{#1}}}}}
\ddefloop ABCDEFGHIJKLMNOPQRSTUVWXYZ\ddefloop
\def\ddef#1{\expandafter\def\csname #1til\endcsname{\ensuremath{\widetilde{#1}}}}
\ddefloop ABCDEFGHIJKLMNOPQRSTUVWXYZ\ddefloop
\def\ddef#1{\expandafter\def\csname tc#1\endcsname{\ensuremath{\widetilde{\mathcal{#1}}}}}
\ddefloop ABCDEFGHIJKLMNOPQRSTUVWXYZ\ddefloop
\def\ddef#1{\expandafter\def\csname #1Bar\endcsname{\ensuremath{\bar{#1}}}}
\ddefloop ABCDEFGHIJKLMNOPQRSTUVWXYZ\ddefloop

\newacronym{ldm}{LDM}{latent diffusion model}
\newacronym{vae}{VAE}{variational autoencoder}
\newacronym{sdvae}{SD-VAE}{Stable Diffusion VAE}

\newcommand{\FigPipeline}{
    \begin{figure}[h]
        \centering
        \includegraphics[width=\textwidth]{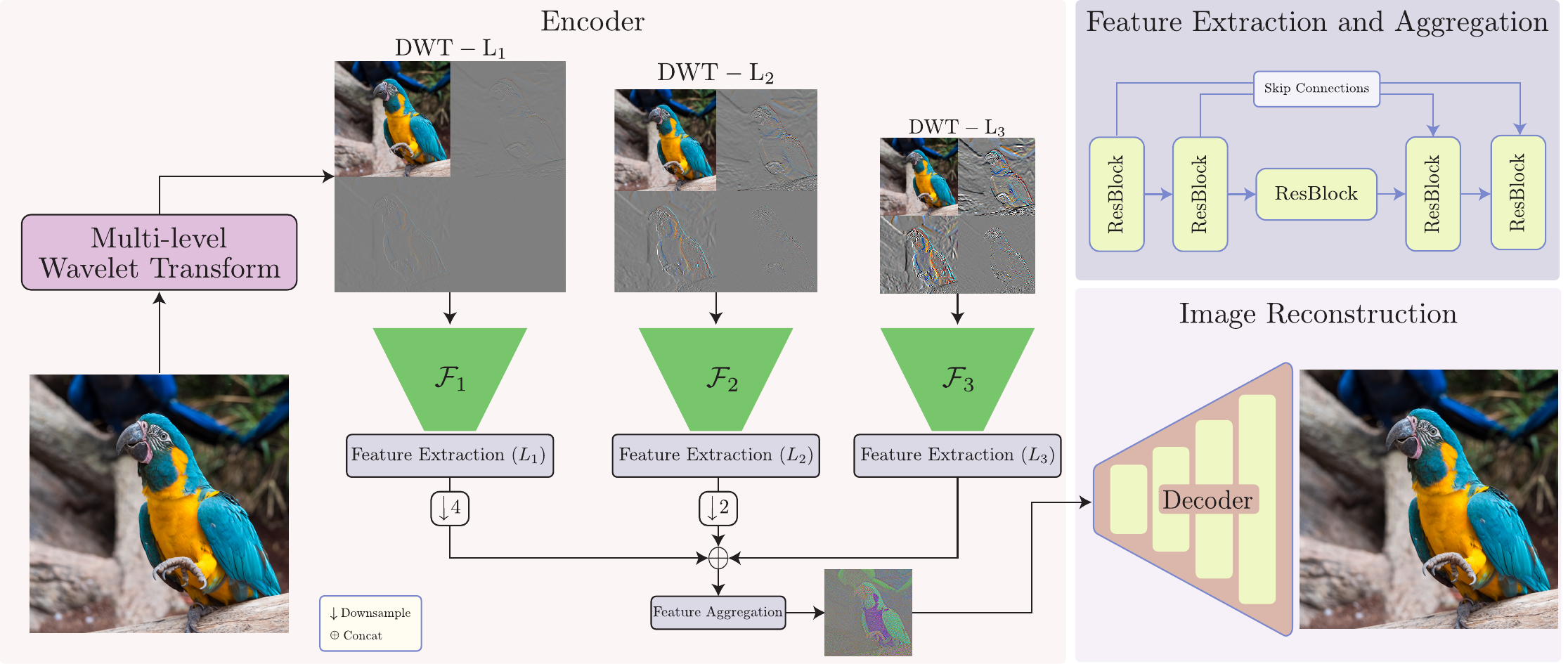}
        \caption{An overview of \method. The input image is first decomposed into multi-level wavelet coefficients, and each wavelet sub-band is separately processed via a feature-extraction network. The features are then combined via a feature-aggregation module to compute the final latent code, which is then transformed back into the image space by the decoder. We use a lightweight UNet architecture (top right) without spatial down/upsampling for feature extraction and aggregation. The decoder is a fully convolutional network similar to that in the Stable Diffusion VAE \citep{rombachHighResolutionImageSynthesis2022}. {\method}'s design allows it to be significantly more efficient than standard VAEs in LDMs while maintaining high reconstruction quality.}
        \label{fig:pipeline}
    \end{figure}
}

\newcommand{\figMain}{
    \begin{figure}[t]
        \centering
        \begin{minipage}{0.8\textwidth}
            \begin{subfigure}[b]{0.32\textwidth}
                \includegraphics[width=\textwidth]{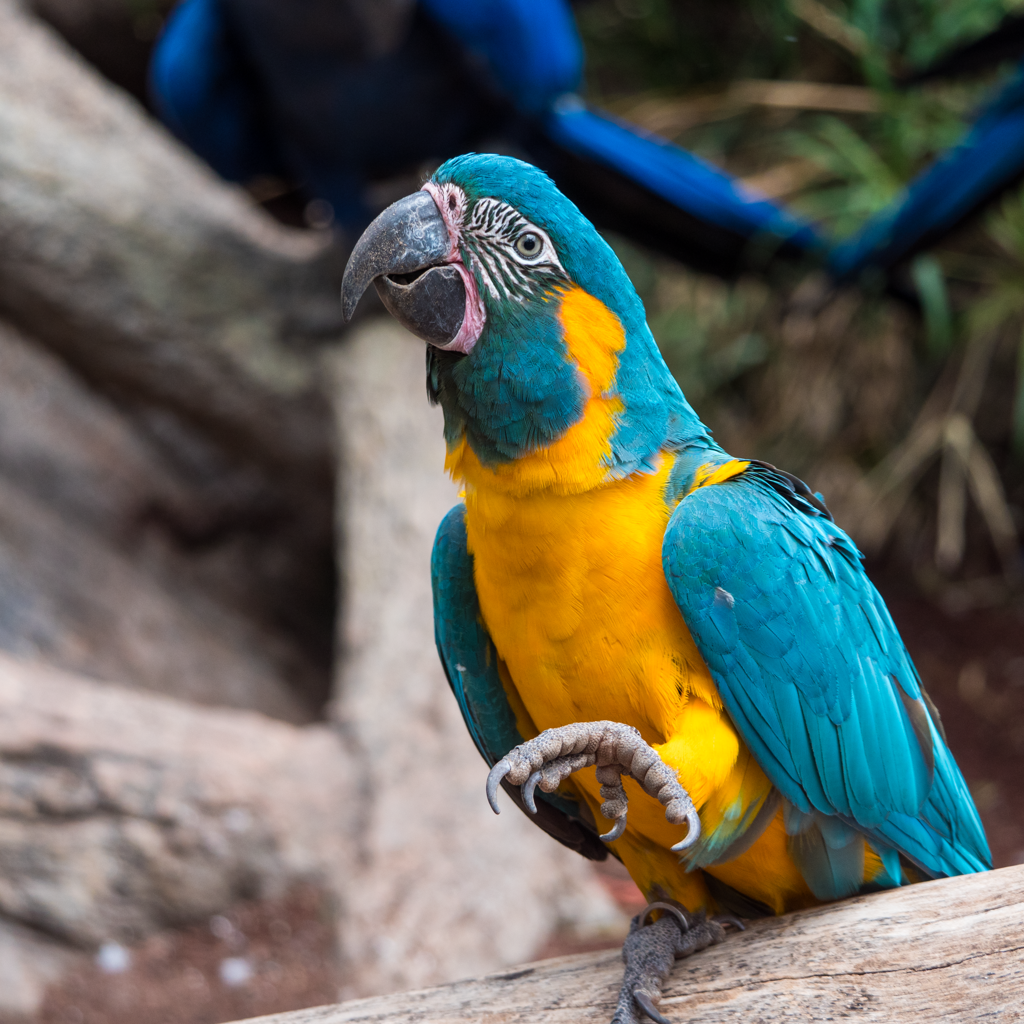}
                \caption{Input}
            \end{subfigure}
            \begin{subfigure}[b]{0.32\textwidth}
                \includegraphics[width=\textwidth]{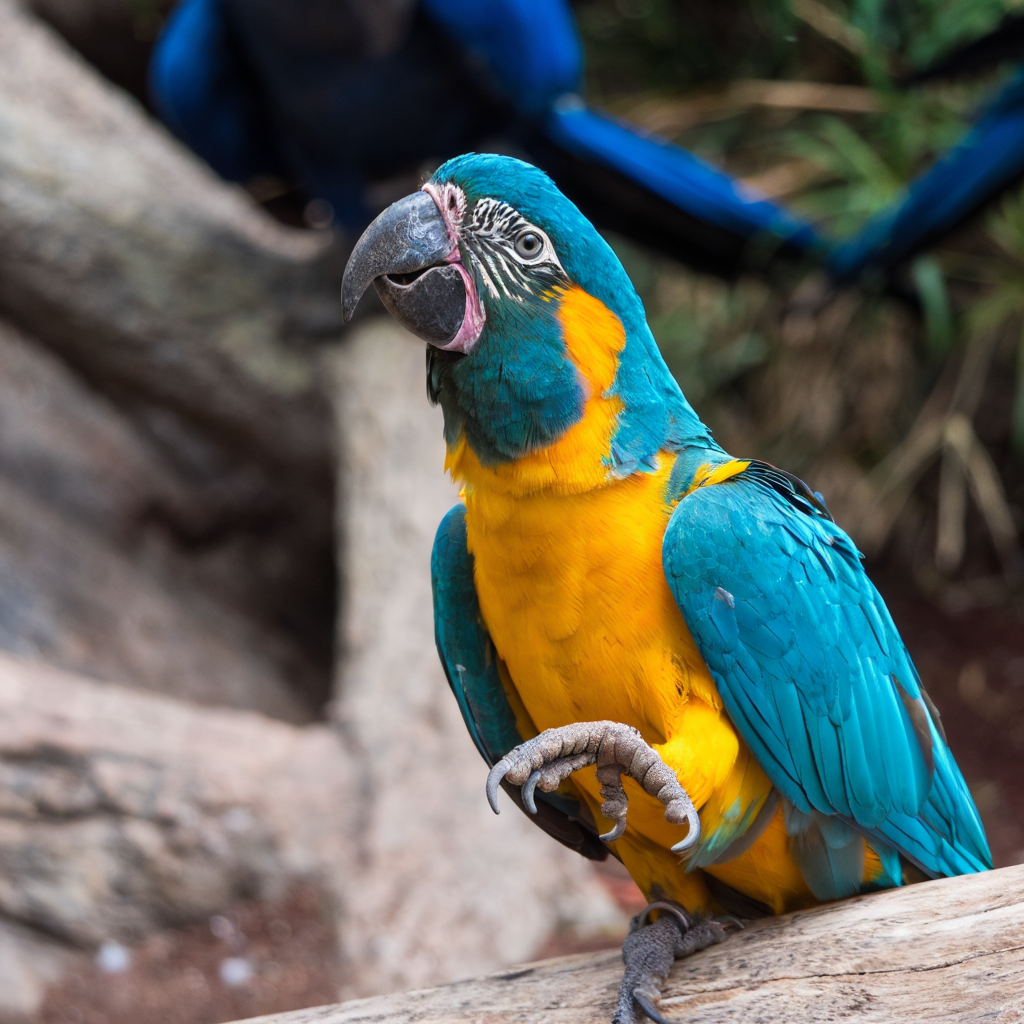}
                \caption{Reconstruction}
            \end{subfigure}
            \begin{subfigure}[b]{0.32\textwidth}
                \includegraphics[width=\textwidth]{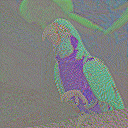}
                \caption{Latent}
            \end{subfigure}
        \end{minipage}
        \caption{An example of the autoencoder reconstruction alongside the learned latent code by \method. We observe that {\method} maintains the image-like structure of SD-VAE.}
        \label{fig:main}
    \end{figure}
}

\newcommand{\FigFeatureMaps}{
\begin{figure}[!t]
    \centering
    \begin{subfigure}[b]{0.24\textwidth}
        \includegraphics[width=\linewidth]{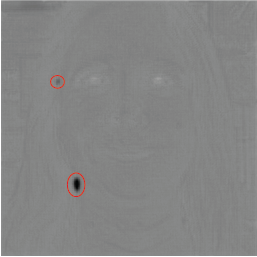}
        \caption*{Group Norm}
    \end{subfigure}
    \begin{subfigure}[b]{0.24\textwidth}
        \includegraphics[width=\linewidth]{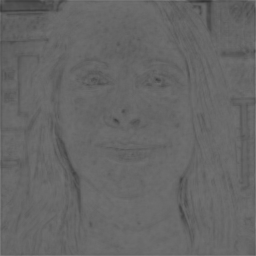}
        \caption*{SMC}
    \end{subfigure}
    \begin{subfigure}[b]{0.24\textwidth}
        \includegraphics[width=\linewidth]{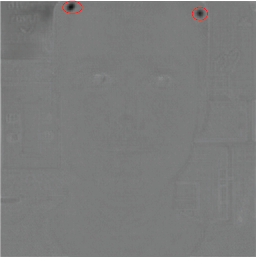}
        \caption*{Group Norm}
    \end{subfigure}
    \begin{subfigure}[b]{0.24\textwidth}
        \includegraphics[width=\linewidth]{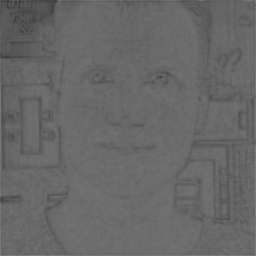}
        \caption*{SMC}
    \end{subfigure}
    \caption{Two examples of the feature maps from the final block of the decoder before and after removing group normalization layers. Using SMC blocks instead of group normalization allows the model to learn more balanced feature maps. The image is best viewed when zoomed in.}
    \label{fig:feature-maps}
    
\end{figure}
}

\newcommand{\figGenerations}{
\begin{figure}[t]
    \centering
    \begin{adjustbox}{valign=b}
    \begin{minipage}{0.56\textwidth}
        \includegraphics[width=\linewidth]{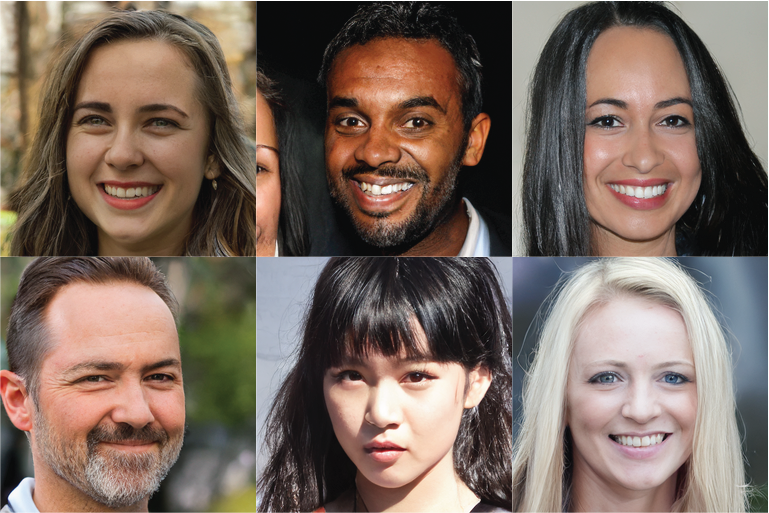}
    \caption[short]{Generated samples from the FFHQ model.}
    \label{fig:generation}
    \end{minipage}
    \end{adjustbox}
    \hfill
    \begin{adjustbox}{valign=b}
        \begin{minipage}{0.40\textwidth}
            \maxsizebox{\linewidth}{!}{
                \begin{tikzpicture}
                    \begin{axis}[
                            xlabel={Training Steps ($\times 10^3$)},
                            ymajorgrids=true,
                            xmajorgrids=true,
                            grid style=dashed,
                            major grid style = {lightgray},
                            ytick style={draw=none},
                            tick label style={font=\small},
                            label style={font=\small},
                            title style={font=\small},
                            legend cell align={left},
                            legend style={font=\small, at={(0.96, 0.54)}},
                            ymode=log,
                        ]

                        \addplot [C0, mark=*, very thick, mark options={solid}] table [x index=0, y index=1, col sep=comma] {data/dweight.csv};
                        \addplot [C1, mark=*, very thick, dashed, mark options={scale=1,solid}] table [x index=0, y index=2, col sep=comma] {data/dweight.csv};
                        \legend{Constant, Adaptive}
    
                    \end{axis}
    
                \end{tikzpicture}
            }
        \caption{Relative gradient norm of the adversarial and the reconstruction loss.}
        \label{fig:dweight}
        \end{minipage}
    \end{adjustbox}
\end{figure}
}

\newcommand{\figScaleDependency}{
    \begin{figure}[t!]
        \centering
        \maxsizebox{\linewidth}{!}{
            \begin{tikzpicture}
                \begin{groupplot}[
                        group style={
                                group size=4 by 1,
                                horizontal sep=1.25cm,
                            },
                        xlabel={Resolution ($\log_2$)},
                        ymajorgrids=true,
                        xmajorgrids=true,
                        grid style=dashed,
                        major grid style = {lightgray},
                        tick label style={font=\normalsize},
                        label style={font=\large},
                        title style={font=\large},
                        legend pos={north east}, legend cell align={left},
                        legend style={font=\small},
                        xtick=data,
                    ]

                    \nextgroupplot[title={rFID}]
                    \addplot [C0, mark=*, very thick, mark options={solid}] table [x index=0, y index=4, col sep=space] {data/wae_scale.dat};
                    \addplot [C1, mark=*, very thick, dashed,  mark options={solid}] table [x index=0, y index=4, col sep=space] {data/vae_scale.dat};
                    \nextgroupplot[title={PSNR}]
                    \addplot [C0, mark=*, very thick, mark options={solid}] table [x index=0, y index=1, col sep=space] {data/wae_scale.dat};
                    \addplot [C1, mark=*, very thick, dashed,  mark options={solid}] table [x index=0, y index=1, col sep=space] {data/vae_scale.dat};
                    \nextgroupplot[title={SSIM}]
                    \addplot [C0, mark=*, very thick, mark options={solid}] table [x index=0, y index=2, col sep=space] {data/wae_scale.dat};
                    \addplot [C1, mark=*, very thick, dashed,  mark options={solid}] table [x index=0, y index=2, col sep=space] {data/vae_scale.dat};
                    \nextgroupplot[title={LPIPS}]
                    \addplot [C0, mark=*, very thick, mark options={solid}] table [x index=0, y index=3, col sep=space] {data/wae_scale.dat};
                    \addplot [C1, mark=*, very thick, dashed,  mark options={solid}] table [x index=0, y index=3, col sep=space] {data/vae_scale.dat};
                    \legend{\method, VAE}

                \end{groupplot}

            \end{tikzpicture}
        }
        \caption{Comparing the performance of {\method} with a normal VAE across different resolutions. {\method} shows less degradation in all metrics.}
        \label{fig:scale-exp}
    \end{figure}

}

\newcommand{\figScaleDependencyModConv}{
    \begin{figure}[t!]
        \centering
        \maxsizebox{\linewidth}{!}{
            \begin{tikzpicture}
                \begin{groupplot}[
                        group style={
                                group size=4 by 1,
                                horizontal sep=1.25cm,
                            },
                        xlabel={Resolution ($\log_2$)},
                        ymajorgrids=true,
                        xmajorgrids=true,
                        grid style=dashed,
                        major grid style = {lightgray},
                        tick label style={font=\small},
                        label style={font=\small},
                        title style={font=\small},
                        legend pos={north east}, legend cell align={left},
                        legend style={font=\small},
                        xtick=data,
                    ]

                    \nextgroupplot[title={rFID}]
                    \addplot [C0, mark=*, very thick, mark options={solid}] table [x index=0, y index=4, col sep=space] {data/wae_scale_modconv.dat};
                    \addplot [C1, mark=*, very thick, dashed,  mark options={solid}] table [x index=0, y index=4, col sep=space] {data/wae_scale_gn.dat};
                    \nextgroupplot[title={PSNR}]
                    \addplot [C0, mark=*, very thick, mark options={solid}] table [x index=0, y index=1, col sep=space] {data/wae_scale_modconv.dat};
                    \addplot [C1, mark=*, very thick, dashed,  mark options={solid}] table [x index=0, y index=1, col sep=space] {data/wae_scale_gn.dat};
                    \nextgroupplot[title={SSIM}]
                    \addplot [C0, mark=*, very thick, mark options={solid}] table [x index=0, y index=2, col sep=space] {data/wae_scale_modconv.dat};
                    \addplot [C1, mark=*, very thick, dashed,  mark options={solid}] table [x index=0, y index=2, col sep=space] {data/wae_scale_gn.dat};
                    \nextgroupplot[title={LPIPS}]
                    \addplot [C0, mark=*, very thick, mark options={solid}] table [x index=0, y index=3, col sep=space] {data/wae_scale_modconv.dat};
                    \addplot [C1, mark=*, very thick, dashed,  mark options={solid}] table [x index=0, y index=3, col sep=space] {data/wae_scale_gn.dat};
                    \legend{with SMC, with Group Norm}

                \end{groupplot}

            \end{tikzpicture}
        }
        \caption{Comparing the performace of {\method} with and without Group Normalization. Using SMC instead of Group Norm makes the autoencoder less scale-dependent.}
        \label{fig:scale-exp-modconv}
    \end{figure}

}

\newcommand{\tabMain}{
\begin{table*}[t]
    \centering
    \caption{Comparison between {\method} and VAE in terms of reconstruction quality across different datasets and latent dimensions. {\method} achieves better or similar reconstruction quality while having considerably fewer parameters in the encoder (34.16M for the VAE and 6.75M for \method). All models use a downscaling factor of \(f=8\) and are trained from scratch with similar training configs (including the choice of loss functions and discriminator). }
    \label{tab:main}
        \small
        \begin{booktabs}{
            colspec = {Q[l, m]Q[c, m]Q[c, m]Q[c, m]Q[c, m]Q[c, m]Q[c, m]}, 
            cell{3,5, 7, 9, 11, 13}{3-Z} = {m, LightGray}, 
            cell{2,4}{1-2} = {r=2}{m},
            cell{6}{1} = {r=4}{m},
            cell{10}{1} = {r=4}{m},
            cell{6,8, 10, 12}{2} = {r=2}{m},
            }
        \toprule
        Dataset & Latent dim & Model & rFID $\downarrow$ & LPIPS $\downarrow$ & PSNR $\uparrow$ & SSIM $\uparrow$ \\
        \midrule
        FFHQ 128 & 16$\times$16$\times$4 & VAE & 0.88 & 0.089 & 28.08 & \textbf{0.85}  \\
         & & \method & \textbf{0.74} & \textbf{0.085} & \textbf{28.36} & \textbf{0.85} \\
         \midrule
          FFHQ 256 & 32$\times$32$\times$4 & VAE & 0.47 & \textbf{0.109} & 28.16 & 0.81  \\
         & & \method & \textbf{0.41} & 0.117 & \textbf{28.33} & \textbf{0.82}\\
         \midrule
         ImageNet 128 & 16$\times$16$\times$4 & VAE & 4.54 & \textbf{0.164} & 24.25 & 0.69  \\
         & & \method & \textbf{4.40} & \textbf{0.164} & \textbf{24.49} & \textbf{0.71} \\
         \cmidrule[l]{2-Z}
            &  16$\times$16$\times$12 & VAE & \textbf{0.94} & \textbf{0.069} & 29.25 & 0.86\\
         & & \method & \textbf{0.94} & \textbf{0.069} & \textbf{29.45} & \textbf{0.87} \\
         \midrule
         ImageNet 256 &  32$\times$32$\times$4 & VAE & 0.89 & 0.160 & 25.83 & 0.73    \\
         & & \method & \textbf{0.87} & \textbf{0.157} & \textbf{26.02} & \textbf{0.74}  \\
         \cmidrule[l]{2-Z}
         & 32$\times$32$\times$12 & VAE & \textbf{0.23} & 0.073 & 30.41 & 0.86	 \\
         & & \method &  \textbf{0.23} & \textbf{0.072} & \textbf{30.91} &	\textbf{0.88}  \\
        \bottomrule
        \end{booktabs}
    \end{table*}
}

\newcommand{\tabDiscriminators}{
    \begin{table}[t!]
        
    \centering
    \caption{Reconstruction quality for different discriminators.}
    \begin{tabular}{lcccc}
        \toprule
        Discriminator & rFID $\downarrow$ & LPIPS $\downarrow$ & PSNR $\uparrow$ & SSIM $\uparrow$  \\ 
        \midrule
        UNet       & \textbf{1.01} & 0.070 & 29.24 & 0.86 \\
        StyleGAN   & 1.38 & 0.065 & 29.51 & \textbf{0.87} \\
        PatchGAN      & 1.61 & \textbf{0.063} & \textbf{29.61} & \textbf{0.87} \\
        \bottomrule
    \end{tabular}
    \label{tab:disc-comparison}
    \end{table}
}

\newcommand{\tabDiscConstantWeight}{
    \begin{table}[t]
    \centering
    \begin{adjustbox}{valign=t}
    \begin{minipage}{0.49\textwidth}
    \centering
        \caption{Reconstruction quality after using constant weight for the adversarial loss.}
        \maxsizebox{\linewidth}{!}{
    \begin{tabular}{lcccc}
        \toprule
        Weight Type       & rFID $\downarrow$ & LPIPS $\downarrow$ & PSNR $\uparrow$ & SSIM $\uparrow$  \\ 
        \midrule
        Adaptive       & \textbf{1.01} & \textbf{0.07} & 29.24 & 0.86 \\
        Constant   & \textbf{1.01} & \textbf{0.07} & \textbf{29.33} & \textbf{0.87} \\
        \bottomrule
    \end{tabular}
    }
    \label{tab:disc-constant-weight}
    \end{minipage}
\end{adjustbox}
\hfill
\begin{adjustbox}{valign=t}
    \begin{minipage}{0.49\textwidth}
    \centering
        \centering
    \caption{Effect of using Gaussian and wavelet loss on final reconstruction quality.}
    \maxsizebox{\linewidth}{!}{
    \label{tab:loss-ablation}
        \begin{tabular}{lcccc}
        \toprule
        Training Config & rFID $\downarrow$ & LPIPS $\downarrow$ & PSNR $\uparrow$ & SSIM $\uparrow$ \\
        \midrule
        Baseline & 0.99 & 0.070 & 29.33 & 0.86 \\
        + Gaussian loss & 0.99 & 0.070 & 29.64 & 0.87 \\
        + wavelet loss & \textbf{0.96} & \textbf{0.069} & \textbf{29.73} & \textbf{0.88} \\
        \bottomrule
        \end{tabular}
    }
    \end{minipage}
\end{adjustbox}
    \end{table}
}

\newcommand{\tabSMSandTrainingRes}{
    \begin{table}[t]
        \begin{adjustbox}{valign=t}
            \begin{minipage}{0.47\textwidth}
                    \centering
                    \caption{Effect of replacing group normalization with SMC on reconstruction quality based on the ImageNet 128$\times$128 model.}
                    \maxsizebox{\columnwidth}{!}{
                    \Large
                    \begin{booktabs}{colspec={lcccc}, row{3}={LightGray}}
                        \toprule
                        Normalization & rFID $\downarrow$ & LPIPS $\downarrow$ & PSNR $\uparrow$ & SSIM $\uparrow$  \\ 
                        \midrule
                        Group Norm       & 1.01 & \textbf{0.07} & 29.24 & 0.86 \\
                        SMC   & \textbf{0.97} & \textbf{0.07} & \textbf{29.32} & \textbf{0.87} \\
                        \bottomrule
                    \end{booktabs}
                    }
                    \label{tab:smc-ablation}
            \end{minipage}
        \end{adjustbox}
        \hfill
        \begin{adjustbox}{valign=t}
            \begin{minipage}{0.50\textwidth}
                \centering
    \caption{Effect of pretraining the autoencoder at lower resolutions. We observe that training at 128$\times$128 followed by fine-tuning at 256$\times$256 performs best.}
    \maxsizebox{\columnwidth}{!}{
    \label{tab:train-res}
        \Large
        \begin{tabular}{lcccc}
        \toprule
        Training Config & rFID $\downarrow$ & LPIPS $\downarrow$ & PSNR $\uparrow$ & SSIM $\uparrow$ \\
        \midrule
        256-full & 0.75 & 0.153 & 26.10 & 0.73 \\
        128-full & 0.97 & 0.162 & 25.90 & 0.72 \\
        128-tuned & \textbf{0.73} & \textbf{0.147} & \textbf{26.22} & \textbf{0.74} \\
        \bottomrule
        \end{tabular}
    }
            \end{minipage}
        \end{adjustbox}
        
    \end{table}
    
}

\newcommand{\tabModelComplexity}{
    \begin{table}[t]
        \centering
        \caption{Comparing the complexity of our encoder with the encoder from the Stable Diffusion VAE for a batch size of 32. The values are measured on one Quadro RTX 6000.}
        \label{tab:model-complexity}
        \maxsizebox{\columnwidth}{!}{
        \begin{booktabs}{colspec={Q[l, m]Q[c, m]Q[c, m]Q[c, m]}, row{3-Z}={LightGray}}
            \toprule
            Encoder & Params (M) & GPU Memory (MB) & Throughput (img/sec) \\
            \midrule
            VAE & 34.16 & 8860  & 68  \\
            \method-S & 1.03 & 1324 & 384  \\
            \method-B & 6.75 & 3155 & 129  \\
            \method-M & 32.75 & 12130 & 42.24  \\
            \method-L & 41.425 & 12130 & 41.6  \\

            \bottomrule
            \end{booktabs}
        }
    \end{table}
}

\newcommand{\tabDiffusionComp}{
    \begin{table}[t!]
        \begin{adjustbox}{valign = t}
            \begin{minipage}{0.49\textwidth}
                \centering
        \caption{Comparing MMD between {\method} latent space and a standard Gaussian vs \gls{sdvae} latent space for different RBF kernels. {\method} is statistically closer to a standard Gaussian.}
        \label{tab:diff-mmd}
        \maxsizebox{\linewidth}{!}{
        \begin{booktabs}{colspec = {lcc}, cell{2-Z}{3}={LightGray}}
            \toprule
            $\sigma$ & \gls{sdvae} & \method \\
            \midrule
            25 & ~~8.67$\pm$0.10 & \textbf{1.44$\pm$0.28} \\
            50 & 28.90$\pm$0.49 & \textbf{7.94$\pm$0.19} \\
            100 & 10.77$\pm$0.29 & \textbf{5.14$\pm$0.19} \\
            250 & ~~1.78$\pm$0.06 & \textbf{1.09$\pm$0.04} \\
            500 & ~~0.44$\pm$0.02 & \textbf{0.28$\pm$0.01} \\
            \bottomrule
            \end{booktabs}
        }
            \end{minipage}
        \end{adjustbox}
        \hfill
        \begin{adjustbox}{valign = t}
            \begin{minipage}{0.49\textwidth}
                \centering
        \caption{Comparison between diffusion models trained in the latent space of a standard VAE \citep{rombachHighResolutionImageSynthesis2022} vs the latent space of {\method}. We observe that both models perform similarly in terms of generation quality.}
        \label{tab:diff-main}
        \maxsizebox{\linewidth}{!}{
        \begin{booktabs}{colspec = {lcccc}, cell{3,5}{2-Z}={LightGray}}
            \toprule
            Dataset & Encoder & FID $\downarrow$ \\
            \midrule
             \SetCell[r=2]{m} FFHQ \citep{stylegan1} (256$\times$256) & LDM & 8.11\\
              & \method  &  \textbf{8.03}\\
              \midrule
              \SetCell[r=2]{m} CelebA-HQ \citep{proggan} (256$\times$256) & LDM & 5.92\\
              & \method  &  \textbf{5.73}\\
            \bottomrule
            \end{booktabs}
        }
            \end{minipage}
        \end{adjustbox}
        
    \end{table}
}

\newcommand{\tabMergeLevel}{
    \begin{table}[t!]
        \centering
        \caption{Ablation on removing the highest resolution wavelets from feature extraction.}
        \label{tab:merge-levels}
        \maxsizebox{\columnwidth}{!}{
        \begin{booktabs}{colspec = {lcccc}}
            \toprule
            Config & rFID $\downarrow$ & LPIPS $\downarrow$ & PSNR $\uparrow$ & SSIM $\uparrow$\\
            \midrule
            All sub-bands & \textbf{0.87} & \textbf{0.157} & \textbf{26.02} & \textbf{0.74}  \\
            Last two sub-bands & 1.20 & 0.17 & 26.04 & 0.74 \\
            \bottomrule
            \end{booktabs}
        }
    \end{table}
}

\newcommand{\tabSharedUNet}{
    \begin{table}[t!]
        \centering
        \caption{Ablation on parameter sharing for the feature-extraction module.}
        \label{tab:shared-unet}
        \begin{booktabs}{colspec = {ccccc}}
            \toprule
            Shared UNet & rFID $\downarrow$ & LPIPS $\downarrow$ & PSNR $\uparrow$ & SSIM $\uparrow$\\
            \midrule
            \xmark &  \textbf{0.75} & \textbf{0.153} & \textbf{26.10} & \textbf{0.73} \\
            \cmark & 0.78 & 0.154 & 26.05 & \textbf{0.73} \\
            \bottomrule
            \end{booktabs}
    \end{table}
}

\newcommand{\tabScaleVAE}{
\begin{table}[t!]
    \vspace*{-0.5cm}
    \centering
    \caption{
        Comparison of the scalability of {\method} with a standard VAE across different model sizes. (a) {\method} matches the performance of the VAE with significantly fewer parameters and outperforms VAEs of similar complexity. (b) A na\"{i}ve downscaling of the VAE performs worse than {\method}. All models use the same decoder. More architecture details are provided in \Cref{sec:imp-detail}.}
        \label{tab:scaling-exp}
    \begin{subtable}{0.49\textwidth}
        \centering
        \caption{Scaling {\method} ($n_z = 12$ for all models)}
        \maxsizebox{\linewidth}{!}{
        \huge
        \begin{booktabs}{colspec={lccccc}, row{3-Z}={LightGray}}
        \toprule
        Model & Params (M) & rFID $\downarrow$ & LPIPS $\downarrow$ & PSNR $\uparrow$ & SSIM $\uparrow$ \\
        \midrule
        VAE & 34.16 & 0.95 & 0.069 & 29.25 & 0.86 \\ 
        \method-S & 1.03 & 1.11 & 0.075 & 29.12 & 0.86 \\
        \method-B & 6.75 & 0.94 & 0.069 & 29.55 & 0.87 \\
        \method-M & 32.75 & 0.79 &  0.064 & 29.68 &  0.87 \\
        \method-L & 41.42 & \textbf{0.74} & \textbf{0.062} & \textbf{29.94} & \textbf{0.88} \\
        \bottomrule
        \end{booktabs}
    }
    \end{subtable}
    \hfill
    \begin{subtable}{0.49\textwidth}
        \centering
        \caption{Downscaling the VAE ($n_z = 4$ for all models)}
    \label{tab:small-vae-comp}
    \maxsizebox{\linewidth}{!}{
        \huge
        \begin{adjustbox}{valign=t}
            \begin{booktabs}{colspec={lccccc}, row{Z}={LightGray}}
                \toprule
                Model & Params (M) & rFID $\downarrow$ & LPIPS $\downarrow$ & PSNR $\uparrow$ & SSIM $\uparrow$\\
                \midrule
                VAE & 34.16 & 4.54 & \textbf{0.164} & 24.25 & 0.69 \\
                VAE-Small & 6.75 & 5.27 & 0.175 & 24.10 & 0.69 \\
                \method-B & 6.75 & \textbf{4.40} & \textbf{0.164} &\textbf{ 24.49} & \textbf{0.71} \\
        \bottomrule
            \end{booktabs}
        \end{adjustbox}
    }
    \end{subtable}
    \vspace*{-0.5cm}
\end{table}
}

\newcommand{\tabVGG}{
    \begin{table}[t!]
        \centering
        \caption{Ablation on using different VGG loss functions for the perceptual loss.}
        \label{tab:vgg-ablation}
        \begin{booktabs}{colspec={lcccc}}
            \toprule
            Perceptual Loss Type & rFID $\downarrow$ & LPIPS $\downarrow$ & PSNR $\uparrow$ & SSIM $\uparrow$\\
            \midrule
            LPIPS \citep{lpips} & 0.99 & \textbf{0.071} & \textbf{29.33} & \textbf{0.86}   \\
            ESRGAN \citep{wang2018esrgan} & \textbf{0.78} & \textbf{0.071} & 28.53 & 0.84 \\
    \bottomrule
        \end{booktabs}
    \end{table}
}

\newcommand{\tabPreConvAblation}{
    \begin{table}[t!]
        \centering
        \caption{Ablation on the effect of 1$\times$1 convolution layers.}
        \label{tab:preconv-ablation}
        \small
        \begin{booktabs}{colspec={ccccc}}
            \toprule
            1$\times$1 Conv & rFID $\downarrow$ & LPIPS $\downarrow$ & PSNR $\uparrow$ & SSIM $\uparrow$\\
            \midrule
            \cmark & 2.04 & 0.190 & 25.61 & 0.72   \\
            \xmark & \textbf{0.87} & \textbf{0.157} & \textbf{26.02} & \textbf{0.74} \\
    \bottomrule
        \end{booktabs}
    \end{table}
}

\newcommand{\tabLDLAblation}{
    \begin{table}[t!]
        \centering
        \caption{Ablation on the effect of adding the LDL loss \citep{jie2022LDL}.}
        \label{tab:ldl-ablation}
        \begin{booktabs}{colspec={ccccc}}
            \toprule
            with LDL & rFID $\downarrow$ & LPIPS $\downarrow$ & PSNR $\uparrow$ & SSIM $\uparrow$\\
            \midrule
            \xmark & 0.99 & \textbf{0.071} & \textbf{29.33} & {0.86}   \\
            \cmark & \textbf{0.98} & \textbf{0.071} & \textbf{29.33} & \textbf{0.87} \\
    \bottomrule
        \end{booktabs}
    \end{table}
}

\newcommand{\tabGanLossAblation}{
    \begin{table}[t!]
        \centering
        \caption{Ablation on using different adversarial loss functions.}
        \label{tab:gan-loss-ablation}
        \begin{booktabs}{colspec={lcccc}}
            \toprule
            Adversarial Loss & rFID $\downarrow$ & LPIPS $\downarrow$ & PSNR $\uparrow$ & SSIM $\uparrow$\\
            \midrule
            Hinge & \textbf{0.99} & {0.071} & {29.33} & {0.86}   \\
            Logistic & 1.00 & \textbf{0.068} & \textbf{29.67} & \textbf{0.88} \\
    \bottomrule
        \end{booktabs}
    \end{table}
}

\newcommand{\tabViTAblation}{
\begin{table}[t!]
    \centering
    \caption{Ablation on using ViT for feature aggregation.}
    \label{tab:vit-ablation}
    \begin{booktabs}{colspec={lcccc}}
        \toprule
        $\mathcal{F}_{\textnormal{agg}}$ & Params (M)  & rFID $\downarrow$ & LPIPS $\downarrow$ & PSNR $\uparrow$ & SSIM $\uparrow$\\
        \midrule
        UNet & 1.69 & {0.94} & \textbf{{0.069}} & 29.25 & 0.86   \\
        ViT & 0.84 & \textbf{{0.92}} & {0.070} & \textbf{{29.44}} & \textbf{{0.87}} \\
\bottomrule
    \end{booktabs}
\end{table}
}

\newcommand{\tabNAFAblation}{
\begin{table}[t!]
    \centering
    \caption{Ablation on using NAFNet \citep{chu2022nafssr} for feature extraction.}
    \label{tab:nafnet-ablation}
    \begin{booktabs}{colspec={lcccc}}
        \toprule
        $\mathcal{F}_{l}$  & rFID $\downarrow$ & LPIPS $\downarrow$ & PSNR $\uparrow$ & SSIM $\uparrow$\\
        \midrule
        UNet & 0.94 & \textbf{0.069} & \textbf{29.55} & \textbf{0.87}   \\
        NAFNet & \textbf{0.93} & \textbf{ 0.069} & 29.36 & \textbf{0.87} \\
\bottomrule
    \end{booktabs}
\end{table}
}

\newcommand{\tabTrainingResVAE}{
\begin{table}[t!]
    \centering
    \caption{Effect of pretraining the autoencoder at lower resolutions for a standard VAE model. The 128-tuned model performs similarly to the model trained solely on 256$\times$256 data.}
    \maxsizebox{\columnwidth}{!}{
    \label{tab:train-res-vae}
        \begin{tabular}{lcccc}
        \toprule
        Training Config & rFID $\downarrow$ & LPIPS $\downarrow$ & PSNR $\uparrow$ & SSIM $\uparrow$ \\
        \midrule
        256-full &  \textbf{0.67} & \textbf{0.150} & \textbf{26.01} & \textbf{0.74} \\
        128-full & 0.89 & 0.161 & 25.83 & 0.73  \\
        128-tuned & 0.69 & 0.151 & 25.97 & 0.73 \\
        \bottomrule
        \end{tabular}
    }
    \end{table}
}

\title{LiteVAE: Lightweight and Efficient Variational Autoencoders for Latent Diffusion Models}

\author{Seyedmorteza Sadat\textsuperscript{1}, Jakob Buhmann\textsuperscript{2}, Derek Bradley\textsuperscript{2}, Otmar Hilliges\textsuperscript{1}, Romann M.\ Weber\textsuperscript{2} \\
\textsuperscript{1}ETH Z\"urich, \textsuperscript{2}DisneyResearch\textbar{}Studios\\
\texttt{\{seyedmorteza.sadat,otmar.hilliges\}@inf.ethz.ch} \\
\texttt{\{jakob.buhmann,derek.bradley,romann.weber\}@disneyresearch.com}
}

\begin{document}

\maketitle

\begin{abstract}
   Advances in \glspl{ldm} have revolutionized high-resolution image generation, but the design space of the autoencoder that is central to these systems remains underexplored. In this paper, we introduce {\method}, a new autoencoder design for LDMs, which leverages the 2D discrete wavelet transform to enhance scalability and computational efficiency over standard \glspl{vae} with no sacrifice in output quality. We investigate the training methodologies and the decoder architecture of {\method} and propose several enhancements that improve the training dynamics and reconstruction quality. Our base {\method} model matches the quality of the established \glspl{vae} in current LDMs with a six-fold reduction in encoder parameters, leading to faster training and lower GPU memory requirements, while our larger model outperforms \glspl{vae} of comparable complexity across all evaluated metrics~(rFID, LPIPS, PSNR, and~SSIM). 
\end{abstract}


\FigPipeline

\section{Introduction}
\Acrfullpl{ldm} \citep{rombachHighResolutionImageSynthesis2022} have recently assumed dominance in the field of high-resolution image generation, primarily due to their scalability and training stability over pixel-space diffusion. The training process of \glspl{ldm} involves two separate stages. In the first, an expressive \acrfull{vae} is trained to transform the raw pixels of an image into a more compact latent representation. In the second, a diffusion model is trained on the latent representations of training images. While numerous studies have investigated the scalability and dynamics of the diffusion component in \glspl{ldm} \citep{peeblesScalableDiffusionModels2022,karras2023analyzing}, the autoencoder element has received far less attention.

 The \gls{vae} in LDMs is not only computationally demanding to train but also affects the efficiency of the diffusion training phase due to the resource requirements of querying a large encoder network for computing the latent codes. For example, as the autoencoder operates on high-resolution images, the VAE encoder of Stable Diffusion 2.1 uses $135.59$ GFLOPS compared with $86.37$ for the diffusion UNet.\footnote{Result of processing a single $256 \times 256 \times 3$ image and its corresponding $32 \times 32 \times 4$ latent representation.} This becomes an even greater concern for video diffusion models, as the encoder then needs to provide the latents for a batch of frames instead of a single image \citep{stableVideoDiffusion}. 
 
 A common workaround for this resource burden is to precompute and cache the latent codes for the entire dataset to avoid having to use the autoencoder during diffusion training. However, in addition to its initial overhead, this approach eliminates the possibility of using on-the-fly techniques, such as data augmentation, which have been shown to improve the training and performance of diffusion models \cite{karras2022elucidating}.  Using a large encoder also adds noticeable overhead in applications that are based on pretrained latent diffusion models. For example, when training 3D models through score distillation of 2D LDMs \citep{DreamFusion}, the process necessitates backpropagating gradients through the LDM encoder, which is computationally intensive \citep{lin2023magic3d}.  Beyond the computational aspects, improving the reconstruction quality of the autoencoder also improves the quality of generated images, as the autoencoder provides an upper bound on the generation quality~\citep{sdxl,esser2024scaling}.

With these issues in mind, we investigated improving the efficiency of LDMs through their core VAE component with the goal of preserving overall quality. We show that with the help of the 2D discrete wavelet transform (DWT), we can considerably simplify the encoder network in \glspl{ldm}. This leads to our proposal of {\method}, a new autoencoder design for LDMs, which has superior compute/quality trade-offs compared with standard VAEs. 

{\method} consists of a lightweight feature-extraction module to compute features from the wavelet coefficients and a feature-aggregation module to combine these multiscale features into a unified latent code. A decoder then converts the latent code back to an image. 
An overview of the {\method} pipeline is shown in \Cref{fig:pipeline}.

We chose the wavelet transform due to its proven ability to represent rich, compact image features \citep{mallat1999wavelet}, and we argue that the wavelet decomposition simplifies the encoder's task by facilitating the learning of meaningful features. 
We examine the design space of {\method} in depth and propose several variations on the network architecture and training setup that further boost reconstruction quality and training efficiency.

Through extensive experimentation, we show that {\method} considerably reduces the computational cost of the standard VAE encoder while maintaining the same level of reconstruction quality. In addition, {\method} provides better reconstruction quality when compared with a VAE of comparable complexity. We also perform an analysis on the latent space learned by {\method} and show that it is similar to that of a regular \gls{vae}.

To summarize, our main contributions in this paper are as follows: 
\begin{enumerate*}[label=(\roman*)]
\item We introduce {\method}, a more efficient and lightweight VAE for LDMs with similar reconstruction quality. This leads to faster training of the autoencoder and higher throughput when training latent diffusion models.
\item We explore the design space of {\method} and propose variations that further enhance reconstruction quality and improve its training dynamics.
\item We perform extensive experimental analyses on the design choices and computational efficiency of {\method} and empirically verify its superior compute efficiency compared to a regular VAE.
\end{enumerate*}

\section{Related work}
\paragraph{Diffusion models and LDMs}
Score-based diffusion models \citep{sohl2015deep, DBLP:conf/nips/SongE19,hoDenoisingDiffusionProbabilistic2020,score-sde} are a class of generative models that learn the data distribution by reversing a forward destruction process that gradually adds Gaussian noise to the data. These models have recently achieved state-of-the-art generation performance on a number of diverse tasks, including unconditional and conditional image generation \citep{nichol2021improved,dhariwalDiffusionModelsBeat2021,karras2022elucidating},  text-to-image synthesis \citep{dalle2,saharia2022photorealistic, rombachHighResolutionImageSynthesis2022, balaji2022ediffi,esser2024scaling}, video generation \citep{blattmann2023align,stableVideoDiffusion,gupta2023photorealistic}, image-to-image translation \citep{saharia2022palette,liu20232i2sb}, and audio generation \citep{WaveGrad,DiffWave,huang2023noise2music}. 

While diffusion models were originally proposed for operating in the ambient image space, \citet{rombachHighResolutionImageSynthesis2022} advocated for following the same methodology in the latent space of a frozen, pre-trained VAE. Following this, a number of advancements have been proposed to enhance latent diffusion models, including architecture improvements \citep{peeblesScalableDiffusionModels2022,maskedDiffusionTransformer,karras2023analyzing}, training setups \citep{min-snr,karras2022elucidating}, and sampling techniques \citep{hoClassifierFreeDiffusionGuidance2022,selfAttentionGuidance,sadat2024cads}. In contrast to these proposed methods, our work focuses on the first stage of LDMs and aims at improving the architecture and efficiency of the VAE component.

\citet{asymmetricVQGAN} recently proposed an improved decoder for the Stable Diffusion VAE that better preserves the details of conditional inputs for tasks such as in-painting. In contrast, our focus in this paper is mainly on the efficiency and properties of the \emph{entire} VAE in LDMs, and our method is not restricted to conditional scenarios. \citet{dai2023emu} also introduced FFT features as input to the VAE for better reconstruction quality. However, their work does not address efficiency, and it can be seen as complementary to ours since FFT features can be combined with our DWT approach to further refine the encoder's initial representation.

\paragraph{Wavelet transform}
The wavelet transformation \citep{waveletIntro,DBLP:journals/pami/Mallat89} is a classic spatial-frequency decomposition of a signal that has gained popularity in numerous computer vision tasks, including denoising \citep{chang2000adaptive, mohideen2008image}, image and video compression \citep{shen1999wavelet, jpeg2000, rippel2017real, ma2020end}, super-resolution \citep{deepwavelet, waveletSR}, and image restoration \citep{figueiredo2003algorithm, mwcnn, yu2021wavefill}. More recently, wavelets have been integrated into generative adversarial networks \citep{SwaGAN, freegan} and pixel-space diffusion models for high-resolution image synthesis \citep{hoogeboom2023simple, waveletDiff, scoreWavelet}. Building on these advancements, we investigate the use of DWT to enhance the efficiency and characteristics of VAEs in LDMs, addressing an underexplored area in the literature.
\section{Background}
This section includes a brief overview of deep autoencoders and the wavelet transform. A summary of diffusion models is given in \Cref{sec:diff-overview}.

\paragraph{Deep autoencoders}

Deep autoencoders consist of an encoder network $\mc{E}$ that maps an image to a latent representation and a decoder $\mc{D}$ that reconstructs the data from the latent code. More specifically, given an input image $\vx \in \mathbb{R}^{H \times W \times 3}$, convolutional autoencoders aim to find a latent vector $\mc{E}(\vx) \in \mathbb{R}^{{H}/{f} \times {W}/{f} \times n_z}$ such that $\mc{D}(\mc{E}(\vx)) \approx \vx$, where \(f\) is the spatial downsampling scale and \(n_z\) is the number of latent channels. 

The training of autoencoders mainly consists of a reconstruction loss $\mc{L}_{\textnormal{recon}}(\mc{D}(\mc{E}(\vx)), \vx)$ between the input image and the reconstructed image, and a regularization term $\mc{L}_{\textnormal{reg}}(\mc{E}(\vx))$ on the latents. $\mc{L}_{\textnormal{recon}}$ is typically a combination of $\ell_1$ and perceptual loss \citep{lpips}, and the regularization $\mc{L}_{\textnormal{reg}}$ can be enforced via Kullback–Leibler (KL) divergence \citep{kingma2013auto} relative to a reference distribution, typically the standard Gaussian. The regularization term forces the latent space to have a better structure for other applications, such as generative modeling. Following \citet{DBLP:conf/cvpr/EsserRO21}, it is also common to train a discriminator $D$ with an adversarial loss $\mc{L}_{\textnormal{adv}}$ that differentiates the real images $\vx$ from the reconstructions $\mc{D}(\mc{E}(\vx))$ for more photorealistic outputs. The overall training loss is then equal to 
\begin{equation}
    \mc{L}_{\textnormal{train}} = \mc{L}_{\textnormal{recon}} + \lambda_{\textnormal{reg}} \mc{L}_{\textnormal{reg}} + \lambda_{\textnormal{adv}} \mc{L}_{\textnormal{adv}},
\end{equation}
where the $\lambda$'s are weighting hyperparameters.
\citet{DBLP:conf/cvpr/EsserRO21} also proposed an adaptive weighting strategy for $\lambda_{\textnormal{adv}}$ given by
\begin{equation}\label{eq:adaptive-weight}
    \lambda_{\textnormal{adv}} = \frac{1}{2} \prn{\frac{\norm{\nabla \mathcal{L}_{\text{recon}}}}{\norm{\nabla \mathcal{L}_{\text{adv}}} + \delta}}
\end{equation}
for a small $\delta > 0$ to balance the relative gradient norm of the adversarial loss with that of the reconstruction loss.

\paragraph{Discrete wavelet transform}
Wavelet transforms are a signal processing technique for extracting spatial-frequency information from input data. Wavelets are characterized by a low-pass filter $L$ and a high-pass filter $H$. For 2D signals, four filters are defined via \(L\trp{L}\), \(L\trp{H}\), \(H\trp{L}\), and \(H\trp{H}\). Given an input image \(\vx\), the 2D wavelet transform decomposes \(\vx\) into a low-frequency sub-band \(\vx_{L}\) and three high-frequency sub-bands \(\{\vx_{H}, \vx_{V}, \vx_{D}\}\) capturing horizontal, vertical, and diagonal details. For an image of size \(H \times W\), each wavelet sub-band is of size \(H/2 \times W/2\). Multi-resolution analysis is achievable by iteratively applying the wavelet transform to \(\vx_{L}\) at each level. Wavelet transforms are also invertible, and one can reconstruct the original image \(\vx\) from the sub-bands \(\{\vx_{L}, \vx_{H}, \vx_{V}, \vx_{D}\}\) using the inverse wavelet transform. Additionally, the Fast Wavelet Transform (FWT) \citep{mallat1989theory} enables the computation of wavelet sub-bands with linear complexity relative to the number of pixels in $\vx$. Consistent with the recent literature \citep{SwaGAN,waveletDiff}, we use Haar basis as the wavelet filter.

\section{Method}\label{sec:method}

\begin{wrapfigure}[11]{r}{0.4\textwidth}
    \centering
    \vspace*{-0.5cm}
    \begin{subfigure}[b]{0.18\textwidth}
        \includegraphics[width=\linewidth]{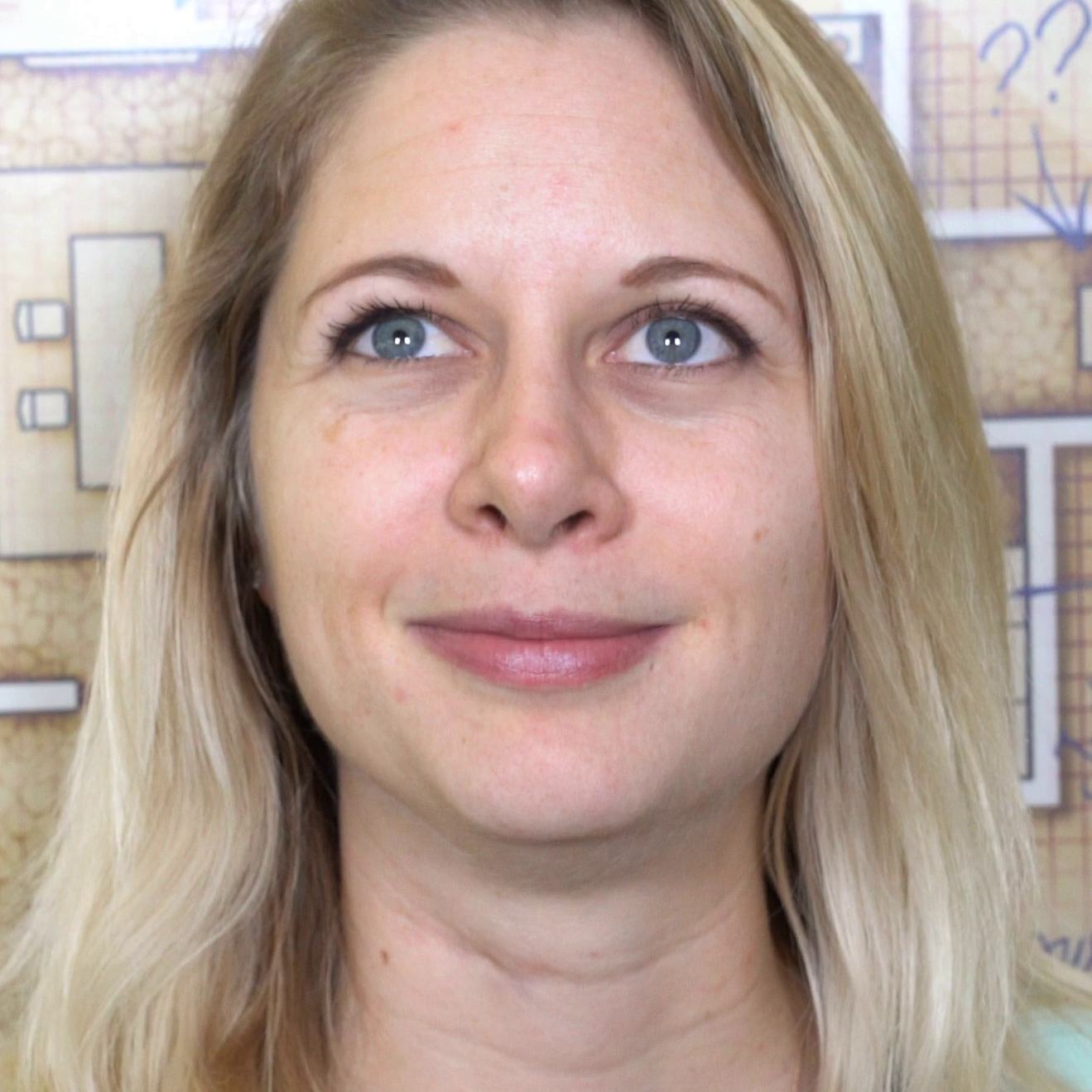}
        \caption{Input image}
    \end{subfigure}
    \begin{subfigure}[b]{0.18\textwidth}
        \includegraphics[width=\linewidth]{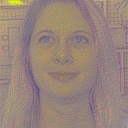}
        \caption{Latent code}
    \end{subfigure}
    \caption{RGB visualization of the first three channels of a SD-VAE latent code.}
    \label{fig:sd-latent}
\end{wrapfigure}
In this section, we describe our design of a more efficient VAE for LDMs and discuss our modifications to the network architectures and the training setup that lead to better reconstruction quality and training efficiency. To motivate our approach, \Cref{fig:sd-latent} shows that when visualizing the latent code learned by the \gls{sdvae}, the code is itself image-like, with a strong similarity to the input. This observation leads us to explore whether the learning of these latent representations can be simplified by applying a fast image-processing function to the input images prior to encoding. We opt for the discrete wavelet transform (DWT) as the image-processing function due to its image-like structure, proven effectiveness in extracting rich, compact features from images, and wide applicability in image-processing tasks such as image compression.

\subsection{Model design}

We now propose {\method}, a wavelet-based autoencoder that reaches the reconstruction quality of standard VAEs with much lower complexity. Our method consists of three main components (see also \Cref{fig:pipeline}):

\textbf{Wavelet processing:} Each image $\vx$  is first processed via a multi-level DWT to get the corresponding wavelet coefficients $\Set{\vx_{L}^{l}, \vx_{H}^{l}, \vx_{V}^{l}, \vx_{D}^{l}}$ at level $l$. To achieve an 8$\times$ downsampling, we use three wavelet levels (i.e., $l \in \Set{1,2,3}$). These features extract multiscale information from $\vx$.   

\textbf{Feature extraction and aggregation:} The wavelet coefficients $\Set{\vx_{L}^{l}, \vx_{H}^{l}, \vx_{V}^{l}, \vx_{D}^{l}}$ are then separately processed via a feature-extraction module $\mathcal{F}_l$ to compute a multiscale set of feature maps  $\mathcal{F}_l(\Set{\vx_{L}^{l}, \vx_{H}^{l}, \vx_{V}^{l}, \vx_{D}^{l}})$. The features are then combined via a feature-aggregation module $\mathcal{F}_{\textnormal{agg}}$ that takes in the output of each $\mathcal{F}_l$ and computes the latent $\vz$. We use a UNet-based architecture similar to the ADM model \citep{dhariwalDiffusionModelsBeat2021} without spatial down/upsampling layers for feature extraction and aggregation. (See \Cref{sec:unlearned} for a discussion of the importance of these learned modules.)

\textbf{Image reconstruction:} Finally, a decoder network $\mathcal{D}$ processes the latent code $\vz$ and computes the reconstructed image $\hat{\vx} = \mathcal{D}(\vz)$.  We use the same decoder network as in \gls{sdvae} for $\mathcal{D}$.

The model is then trained end-to-end to learn the parameters of $\Set{\mathcal{F}_{l}}$, $\mathcal{F}_{\textnormal{agg}}$, and $\mathcal{D}$. Because different wavelet levels already contain enough information about the images, we can use lightweight networks for the feature extraction and aggregation steps. Hence, {\method} essentially combines the computational benefits of DWT with the expressiveness of a learned encoder. Please refer to \Cref{sec:imp-detail,sec:pseudocode} for implementation details.

\subsection{Self-modulated convolution}\label{sec:smc}
In addition to improving the encoder, we observe that the intermediate feature maps learned by the decoder are relatively imbalanced, with certain areas having significantly stronger magnitudes. An example of this issue is shown in \Cref{fig:feature-maps}. Consistent with \citet{stylegan1}, we argue that this issue is due to excessive group normalization layers \citep{wu2018group} in the decoder architectures typically used in autoencoders, since such layers potentially destroy any information found in the magnitudes of the features relative to each other \citep{Karras2019stylegan2}. 

We propose a modified version of modulated convolution \citep{Karras2019stylegan2} instead of group normalization to avoid imbalances. Instead of modulating the convolution layers via a data-dependent style vector, we allow the convolution layer to learn the corresponding scales for each feature map. We call this operation self-modulated convolution (SMC). SMC modifies the convolution weights \(w_{ijk}\) according to
\begin{equation}
    w_{ijk}' = \frac{s_i w_{ijk}}{\sqrt{\sum_{i,k}(s_i w_{ijk})^2} + \epsilon}
\end{equation}
for $\epsilon > 0$, where \(s_i\) is a learnable parameter, and $\{i,j,k\}$ spans the input feature maps, output feature maps, and the spatial dimension of the convolution. Our experiments show that using SMC in the decoder balances the feature maps and also improves the final reconstruction quality due to better training dynamics. Two examples of the decoder feature maps after using SMC are shown in~\Cref{fig:feature-maps}. 
\FigFeatureMaps

\subsection{Training improvements}
Besides the network architecture, we also introduce the following modifications that further enhance the training dynamics and reconstruction quality of {\method}. We verify the effect of these modifications in \Cref{sec:experiments,sec:ablation}.

\paragraph{Training resolution}
While the autoencoders in LDMs are typically trained on 256$\times$256 data (similar to SD-VAE), we observe that the bulk of the training of {\method} can be effectively conducted at a lower 128$\times$128 resolution. Our experiment suggests that pretraining at this lower resolution followed by a fine-tuning stage at the full resolution achieves similar reconstruction quality while requiring significantly less compute for most of the training. We later show in \Cref{sec:vae-res} that this improvement is also generally applicable to the standard VAE models.

\paragraph{Improving the adversarial setup}
 We replace the PatchGAN discriminator used in Stable Diffusion with a UNet-based model for pixel-wise discrimination \citep{schonfeld2020u}. We also notice that the adaptive weight (\Cref{eq:adaptive-weight}) for the adversarial loss update does not introduce any benefit and can be removed for more stable training, especially in mixed-precision setups. 
 
 \paragraph{Additional loss functions} We also introduce two high-frequency reconstruction loss terms based on the wavelet transform and Gaussian blurring \citep{zamfir2023towards}. Let \(\vx\) be the input image and \(\hat{\vx}\) the corresponding reconstruction. For the wavelet term, we compute the  Charbonnier loss \citep{barron2019general} between the high-frequency DWT sub-bands \(\Set{\vx_{H}, \vx_{V}, \vx_{D}}\) and \(\Set{\hat{\vx}_{LH}, \hat{\vx}_{HL}, \hat{\vx}_{HH}}\). For the Gaussian loss, given a Gaussian filter $h$, we compute the \(\ell_1\) loss between \(\vx - h(\vx)\) and \(\hat{\vx} - h(\hat{\vx})\).

\section{Experiments}\label{sec:experiments}
This section presents a comprehensive empirical evaluation of {\method}, demonstrating its superior trade-off between computational efficiency and quality relative to standard VAEs. We further explore the properties of {\method} along with the changes proposed in \Cref{sec:method}. For each experiment, all models in comparison are trained with the exact same training setup, including the loss functions and the discriminator, to ensure a fair comparison.

\paragraph{Evaluation metrics} We follow the same evaluation pipeline as in \citet{rombachHighResolutionImageSynthesis2022} and use reconstruction Fréchet Inception Distance (rFID) \citep{fid} as the main metric to measure the quality and realism of autoencoder outputs due to its alignment with human judgment. For completeness, we also report PSNR, SSIM, and LPIPS \citep{lpips}. As FID is sensitive to small implementation details \citep{cleanFID}, we recompute the metrics as much as possible based on released checkpoints to have a fair comparison between different models.

\paragraph{Main results}
\tabMain

We first demonstrate that {\method} matches or exceeds the performance of standard VAEs across various datasets and latent dimensions, as shown in \Cref{tab:main}. Notably, the model employed for this table utilizes approximately one-sixth of the encoder parameters compared to the VAE model (6.75M vs 34.16M) and hence trains faster. Also, one example of the reconstruction quality and the learned latent representation by {\method} is given in \Cref{fig:main}. We notice that {\method} maintains the image-like latent codes, similar to the SD-VAE latent in \Cref{fig:sd-latent}. 

\figMain

\paragraph{Increasing model complexity}

In \Cref{tab:scaling-exp} we show the scalability of {\method} as we increase the complexity of the feature-extraction and feature-aggregation blocks. We note that the reconstruction performance strictly improves by using more encoder parameters, and our large models outperform a standard VAE of similar complexity across all metrics. Hence, we conclude that {\method} offers superior scalability w.r.t.\ the model size. 

\paragraph{Scaling down the encoder in VAEs}

 \Cref{tab:small-vae-comp} also indicates that the na\"{i}ve approach of scaling down the encoder in standard VAEs does not perform on par with our method in terms of reconstruction quality. Thus, we conclude that {\method} takes better advantage of the encoder parameters than normal VAEs, mainly due to the wavelet processing step that provides the encoder with a rich representation from the beginning.
\tabScaleVAE

\paragraph{Computational cost}
\Cref{tab:model-complexity} presents a comparison of the computational costs between {\method} and the Stable Diffusion VAE encoder. {\method}-B requires considerably less GPU memory and offers nearly double the throughput. This reduction in computational complexity allows the usage of larger batch sizes when training the autoencoder, as shown to be beneficial by \citet{sdxl}, and leads to better hardware utilization for diffusion training in the second stage of LDMs since fewer resources should be devoted to computing the latent input for the diffusion model.
\tabModelComplexity

\tabSMSandTrainingRes
\paragraph{Removing group normalization in the decoder}

We qualitatively showed in \Cref{fig:feature-maps} that group normalization in the decoder causes imbalanced feature maps in the network and that SMC can remove such artifacts. Here we also quantitatively show in \Cref{tab:smc-ablation} that replacing group normalization with SMC leads to better reconstruction quality. Additionally, we demonstrate in \Cref{sec:scale-smc} that removing the imbalanced feature maps results in less scale dependency in the final model.

\paragraph{Training resolution} 

We next demonstrate the feasibility of pretraining {\method} at a lower resolution of 128$\times$128 followed by a fine-tuning step on 256$\times$256 images. To illustrate this, we compare a model trained for 150k steps at full resolution (256-full) with one trained for 100k steps at 128 and an additional 50k steps at 256 (128-tuned). As shown in \Cref{tab:train-res}, the 128-tuned model even slightly outperforms the model fully trained at the higher resolution. We also note that fine-tuning is essential, as the model trained solely on 128$\times$128 images for 150k steps (128-full) performs worse than the other two. This experiment implies that the model can learn most of the semantics at lower resolutions and recover additional higher-frequency contents in the fine-tuning stage. This pretraining technique reduced the overall wall-clock time of our training runs at 256$\times$256 resolution by more than a factor of two.
\paragraph{Scale dependency}
\figScaleDependency

\Cref{fig:scale-exp} demonstrates that compared to the standard VAEs,  {\method} is less prone to performance degradation when evaluating the model at different resolutions. We hypothesize that as our model learns features on top of multi-resolution wavelet coefficients, it is able to learn more scale-independent features compared to a standard encoder and leave the specific details of each scale to the initial wavelet processing step.

\paragraph{Analysis of the {\method} latent space} 
\tabDiffusionComp
\figGenerations
We also analyzed the characteristics of the latent space of {\method}.  Qualitative inspection of Figures \ref{fig:sd-latent} and \ref{fig:main}, which are representative of the results that hold across our data, show that our latent space and \gls{sdvae} share a similar image-like structure. Separately, we also examined the statistical distance between our model's latent space and pure Gaussian noise.  The intuition here is that, since a diffusion model will have to form a path from pure Gaussian noise to our model's latent space, we do not want that path to be longer than the path a diffusion model has to form between Gaussian noise and the Stable Diffusion latent space.  To this end, we compute the maximum mean discrepancy (MMD) \citep{gretton2012kernel} between latent codes from {\method} and samples from a standard Gaussian and compare the result with that observed for the \gls{sdvae}  (See \Cref{tab:diff-mmd}).  Here the MMD serves as a proxy measure for the path length between these distributions.  In all tested cases, over a variety of RBF kernel bandwidths, our latent space is closer to Gaussian noise than that of \gls{sdvae}. 

Lastly, we trained two diffusion models on the FFHQ and CelebA-HQ datasets and compared their performance with standard VAE-based LDMs. The diffusion model architecture used for this experiment is a UNet identical to the original model from \citet{rombachHighResolutionImageSynthesis2022}. \Cref{tab:diff-main} shows that the diffusion models trained in the latent space of {\method} perform similarly to (or slightly better than) the standard LDMs. Additionally, \Cref{fig:generation} includes some generated examples from our FFHQ model. These results suggest that diffusion models are also capable of modeling the latent space of {\method}.

\section{Ablation studies}\label{sec:ablation}
We next present our main ablation studies to determine the individual impact of the changes proposed in \Cref{sec:method}. We use the ImageNet 128$\times$128 model with a latent size of 32$\times$32$\times$12 as the baseline for all ablations. Further ablation studies on other design choices in {\method} are provided in \Cref{sec:ablation-appendix}.

\paragraph{Removing adaptive weight for $\lambda_{\textnormal{reg}}$}
\tabDiscConstantWeight 
\Cref{tab:disc-constant-weight} demonstrates that we can safely remove the adaptive weight for the adversarial loss (\Cref{eq:adaptive-weight}) and still slightly improve the metrics. \Cref{fig:dweight} also shows the relative norm of the gradient of the adversarial loss compared to the reconstruction loss for both adaptive and constant \(\lambda_{\textnormal{adv}}\). We observe that using adaptive \(\lambda_{\textnormal{adv}}\) leads to more imbalanced gradient ratios, and hence less stable training, especially for mixed-precision scenarios.  Accordingly, we exclusively use a constant weight for the adversarial loss in our experiments.

\paragraph{High-frequency loss functions}
\Cref{tab:loss-ablation} shows the effect of adding high-frequency losses based on Gaussian filtering and the wavelet transform. The addition of these high-frequency loss terms during training consistently improves all reconstruction metrics.

\paragraph{Choice of the discriminator}
\tabDiscriminators

We finally show that using a UNet-based discriminator \citep{schonfeld2020u} outperforms both PatchGAN and StyleGAN discriminators used in previous works \citep{rombachHighResolutionImageSynthesis2022,yu2022vectorquantized} in terms of rFID while having comparable performance for other metrics. We also empirically noted that using a UNet discriminator resulted in more stable training across different runs and hyperparameters. The full comparison for this experiment is given in \Cref{tab:disc-comparison}.
\section{Conclusion}\label{sec:conclusion}
In this paper, we presented {\method}, a new design concept for autoencoders based on the multi-resolution wavelet transform.  {\method} can match the performance of standard VAEs while requiring significantly less compute. We also analyzed the design space and training of this proposed family of autoencoders and offered several modifications that further improve the final reconstruction quality and training dynamics of the base model. Overall, {\method} offers more flexibility in terms of performance/compute trade-off and outperforms the na\"{i}ve approach of making the VAE encoder smaller. Our current work is focused on improving efficiency in the models responsible for encoding the latent representation of natural images, and whether the efficiency benefits of LiteVAE extend to other domains is a question we leave to follow-up work. Although we introduced {\method} in the context of \glspl{ldm}, we hypothesize that its application is not confined to this scenario. We consider the extension of {\method} to other autoencoder-based generative modeling schemes (e.g., tokenization) a promising avenue for further research.

{
\bibliographystyle{plainnat}
\bibliography{ref}
}

\clearpage
\appendix
\section{Broader impact statement}\label{sec:impact-statement}
Our work can significantly reduce the training time and memory requirements of autoencoders in latent diffusion models (LDMs). Given the rising popularity of LDMs, our approach holds promise for positive environmental impacts and significant advancements in generative modeling. It is important to note that while AI-generated content can enhance productivity and creativity, we must remain mindful of the potential risks and ethical concerns involved. For a deeper discussion of ethics and creativity in computer vision, readers are directed to \citep{rostamzadeh2021ethics}.
\section{Using a non-learned encoder}\label{sec:unlearned}
This section provides further motivation behind the design of {\method}. We investigate using a non-learned (i.e., fixed) encoder in two settings: (1) for simple datasets such as FFHQ \citep{stylegan1} and (2) for more diverse datasets such as ImageNet \citep{imagenet}. We use the reconstruction FID (rFID) \citep{fid} as our measure of reconstruction quality, aiming to achieve the \gls{sdvae}'s downsampling factor of \(f=8\). The analysis leads to two observations. First, the non-learned autoencoder (although efficient) can provide high-quality reconstructions only if we use a larger channel depth for the encoder network compared to \gls{sdvae}. Secondly, the dense latent space learned by the autoencoder provides a better structure for generative modeling. {\method} essentially combines the computational benefits of the non-learned encoder with the learned latent space of a regular VAE.

\paragraph{Simple datasets} 
\begin{wraptable}[9]{r}{0.325\textwidth}
    \centering
    \vspace{-0.35cm}
    \caption{The performance of the DWT-based encoder on simple datasets.}
        \label{tab:det-enc-simple}
        \maxsizebox{\linewidth}{!}{
            \begin{adjustbox}{valign=t}
            \begin{booktabs}{colspec = {lccccc}}
            \toprule
            Dataset & Encoder & $n_z$ & rFID $\downarrow$ \\
            \midrule
             \SetCell[r=2]{m,l} FFHQ & \gls{sdvae} & 4 & 0.85\\
              & DWT & 12 & 0.70 \\
              \midrule
              \SetCell[r=2]{m,l} DeepFashion & \gls{sdvae} & 4 & 1.64\\
              & DWT & 12 & 1.71 \\
            \bottomrule
            \end{booktabs}
        \end{adjustbox}
        }
\end{wraptable}
For simpler datasets like FFHQ, it is possible to completely replace the encoder \(\mathcal{E}\) with a predefined function and get similar reconstruction quality. In our case, we used a three-level DWT and only kept the sub-bands of the lowest level. We then trained a decoder to convert the lowest-level sub-bands back to the image. \Cref{tab:det-enc-simple} shows the results of this approach on two relatively restricted datasets. We observe that this wavelet representation offers a similar reconstruction quality to a learned encoder while reducing the number of encoder parameters from about 34M to zero. This experiment indicates that with the help of rich image representations from the wavelet transform, we can speed up SD-VAE by reducing the complexity of the encoder.

\paragraph{Complex datasets}
\begin{wraptable}[8]{r}{0.35\textwidth}
    \centering
    \vspace{-0.3cm}
    \caption{The performance of the non-learned encoder on ImageNet.}
        \label{tab:det-enc-comlex}
        \maxsizebox{\linewidth}{!}{
            \begin{adjustbox}{valign=t}
            \begin{booktabs}{colspec = {lccccc}}
                \toprule
                Dataset & Encoder & $n_z$ &rFID $\downarrow$ \\
                \midrule
            \SetCell[r=3]{m} ImageNet & \gls{sdvae} & 4 & 0.80\\
                & DWT & 12 & 1.65 \\
                & DWT-2 & 48 &  0.32 \\
                \bottomrule
            \end{booktabs}
        \end{adjustbox}
        }
\end{wraptable}
The next step is to explore whether this non-learned encoder setup is scalable to more diverse datasets such as ImageNet. \Cref{tab:det-enc-comlex} demonstrates that while the non-learned encoder is effective in simpler scenarios, it falls short of the quality of normal VAEs in more complex settings. This indicates that the information present in the higher frequency sub-bands of the wavelet transform is essential for the decoder to reconstruct more diverse images with higher quality. To validate this hypothesis, we incorporate the information from higher frequency sub-bands via a space-to-depth operation \citep{shi2016real} in the encoder and observe that we can recover the high reconstruction quality of the learned encoder (DWT-2 in \Cref{tab:det-enc-comlex}). However, this approach is not preferable because the channel dimension is now too high for generative modeling.

\paragraph{Importance of having a learned latent space}
\begin{wraptable}{r}{0.4\textwidth}
    \centering
        \caption{Comparison between a non-learned encoder and  {\method} for training diffusion models.}
        \label{tab:diff-comp-dwt}
        \maxsizebox{\linewidth}{!}{
        \begin{booktabs}{colspec = {lcccc}}
            \toprule
            Dataset & Encoder & FID $\downarrow$ \\
            \midrule
             \SetCell[r=2]{m} FFHQ (256$\times$256) & non-learned & 12.51\\
              & \method  &  \textbf{8.03}\\
            \bottomrule
            \end{booktabs}
        }
            \vspace{-0.2cm}
\end{wraptable}
Finally, we demonstrate that although it is possible to completely replace the encoder of the VAE with a non-learned wavelet-based latent representation for the FFHQ dataset, the learned latent space in {\method} offers a better structure for training diffusion models. \Cref{tab:diff-comp-dwt} indicates that training the diffusion model on the learned latent code of {\method} outperforms the non-learned DWT representation. We argue that the sparse nature of wavelets is harmful to generation quality compared to the dense representation learned by the encoder of {\method}.
\section{Summary of diffusion models}\label{sec:diff-overview}
Diffusion models learn the data distribution \(\pdata\) by reversing a noising process that gradually converts a data point \(\vx\) into random Gaussian noise. More specifically, diffusion models define a forward process via \(\vx_t = \vx + \sigma(t) \beps\), where \(\beps \sim \normal{0}{\mI}\). Then, they train a denoiser network \(D_{\bmtheta}\) to estimate the clean signal \(\vx\) from the current noisy sample \(\vx_t\). It has been shown that this process corresponds to the following stochastic differential equation (SDE) \citep{score-sde,karras2022elucidating}
\begin{equation}\label{eq:diffusion-sde}
 \odif{\vx}_t = - \dot{\sigma}(t)\sigma(t)\grad_{\vx_t} \log p_t(\vx_t) \odif{t} - \beta(t) \sigma(t)^2 \grad_{\vx_t} \log p_t(\vx_t) \odif{t} + \sqrt{2 \beta(t) } \sigma(t)\odif{\omega_t},
\end{equation}
where $\odif{\omega_t}$ is the standard Wiener process, $p_t(\vx_t)$ is the distribution of noisy samples at time $t$, and $\beta(t)$ is a term that controls the influence of noise during the sampling process.
The denoiser network \(D_{\bmtheta}\) effectively approximates the score function $\grad_{\vx_t} \log p_t(\vx_t)$. Given that $p_0=\pdata$ and  $p_1=\normal{\zero}{\sigma_{\textnormal{max}}^2\mI}$, sampling new data points is then possible by starting from random Gaussian noise and solving the corresponding SDE reverse in time.

Latent diffusion models \citep{rombachHighResolutionImageSynthesis2022} follow the same methodology, but instead of performing the forward and backward process in the pixel space, they first convert the data into the latent codes via a pretrained \gls{vae} and employ the diffusion process in the latent space. Please refer to \citet{karras2022elucidating} and \citet{yang2022diffusion} for more details on diffusion models. 

\section{Additional ablation studies}\label{sec:ablation-appendix}
This section contains additional ablation studies on the design space and training dynamic of {\method}.

\subsection{Training loss functions}
We experimented with the following changes to the loss functions during training of the autoencoder to measure whether they lead to any improvement in reconstruction quality. 

\paragraph{Changing the VGG loss} 
\citet{wang2018esrgan} proposed a different VGG loss function based on the features \emph{before} the activation layer. We also ablated this choice against the standard LPIPS loss typically used in the VAEs of LDMs. \Cref{tab:vgg-ablation} indicates that this change has a considerable boost to the reconstruction FID at the cost of lower PSNR. As the perceptual quality is more important for  LDMs compared to distortion metrics, we recommend switching to this loss function instead of the LPIPS loss. However, we used the LPIPS loss for the experiments in the main text to have a similar training setup with commonly used VAEs in LDMs.

\paragraph{Including the locally discriminative learning (LDL) loss}
\citet{jie2022LDL} introduced the LDL loss function to reduce the artifacts caused by the discriminator in the super-resolution context. We also experimented with this loss term and found that it does not have any noticeable impact on the reconstruction quality of {\method}, as shown in \Cref{tab:ldl-ablation}. 

\paragraph{Choosing different adversarial loss functions} 
We also ablated the adversarial loss function for two different setups: a hinge loss, and a non-saturating (logistic) loss. As depicted in \Cref{tab:gan-loss-ablation}, we observe that the hinge loss generally leads to slightly better rFID while the logistic loss achieves slightly better PSNR. Since the adversarial loss in the autoencoder training is only responsible for increasing the photorealism of the outputs, we conclude that both loss terms work equally well.


\subsection{Role of the 1$\times$1 convolution layers}
\citet{dalle} showed that using a 1$\times$1 convolution after the output of the encoder and before the input of the decoder improves the approximation accuracy of the evidence lower bound (ELBO) term in the loss function. We ablated this design choice in the context of {\method} and found that restricting the receptive field of the latent space with these 1$\times$1 convolution layers might be harmful to the reconstruction quality by enforcing too much KL regularization. \Cref{tab:preconv-ablation} shows that removing these convolution blocks leads to much better reconstruction quality in our 256$\times$256 model. Accordingly, we suggest removing these 1$\times$1 convolutions from the model (or, equivalently, adjusting the weight of the KL loss) to get better reconstruction.

\tabVGG
\tabLDLAblation
\tabGanLossAblation
\tabPreConvAblation
\tabNAFAblation

\subsection{Different networks for feature extraction}
We also experimented with NAFNet \citep{chu2022nafssr} instead of the UNet for extracting features from wavelet sub-bands and observed that it performs similarly to the UNet architecture mentioned in the main text. The results are given in \Cref{tab:nafnet-ablation}. This experiment indicates that other network choices for the feature-extraction module are indeed possible, and {\method} is flexible w.r.t.\ this design choice. We chose the UNet to keep the setup as close as possible to the standard VAE design in LDMs.

\subsection{Sharing the weights of the feature-extraction UNet} 
We next investigated whether a single UNet could be shared across different wavelet sub-bands to further reduce the encoder’s trainable parameters.  \Cref{tab:shared-unet} demonstrates that it is indeed possible to share $\mathcal{F}_l$ between different sub-bands. A shared UNet might lead to the post hoc usage of the encoder across different wavelet levels and resolutions at inference. However, as the computational cost (in terms of GFLOPS) does not change with parameter sharing, we did not use this technique for the main experiments.

\subsection{Using ViT for feature aggregation} 
We also explore the use of non-convolutional vision transformer (ViT) blocks \citep{vit} for feature aggregation $\mathcal{F}_{\textnormal{agg}}$. As indicated in \Cref{tab:vit-ablation}, employing ViT achieves comparable reconstruction quality to that of a fully-convolutional encoder, but with fewer parameters. However, it is important to note that incorporating ViT makes the model resolution-dependent. This is a drawback, as the VAE in LDMs is usually required to operate on data with varying resolutions. Hence, we side with the UNet models to make the encoder resolution-independent.

\subsection{Importance of using all wavelet levels}
We also explored the possibility of performing feature extraction on only a subset of wavelet coefficients rather than across all wavelet levels. As shown in \Cref{tab:merge-levels}, this approach negatively impacts reconstruction performance on ImageNet, indicating that incorporating information from all wavelet levels is essential for high-quality reconstruction, particularly with complex datasets.

\subsection{Scale dependency of SMC}\label{sec:scale-smc}
This section shows that using SMC improves the scale dependency of {\method}. The results in \Cref{fig:scale-exp-modconv} indicate that using SMC instead of group normalization leads to less degradation in performance as we change the resolution of the evaluation dataset. We argue that removing the imbalanced feature maps aids the network in learning features that are less scale-dependent.

\subsection{Training resolution for the standard VAEs}\label{sec:vae-res}
This experiment validates that the idea of pretraining the autoencoder at 128$\times$128 followed by fine-tuning at 256$\times$256 also works for the standard VAEs. The results of this experiment are given in \Cref{tab:train-res-vae}. Similar to {\method}, the 128-tuned model matches the performance of the 256-full model while requiring considerably less training compute.

\section{Additional generated samples}
\begin{figure}[t!]
    \centering
    \includegraphics[width=0.8\textwidth]{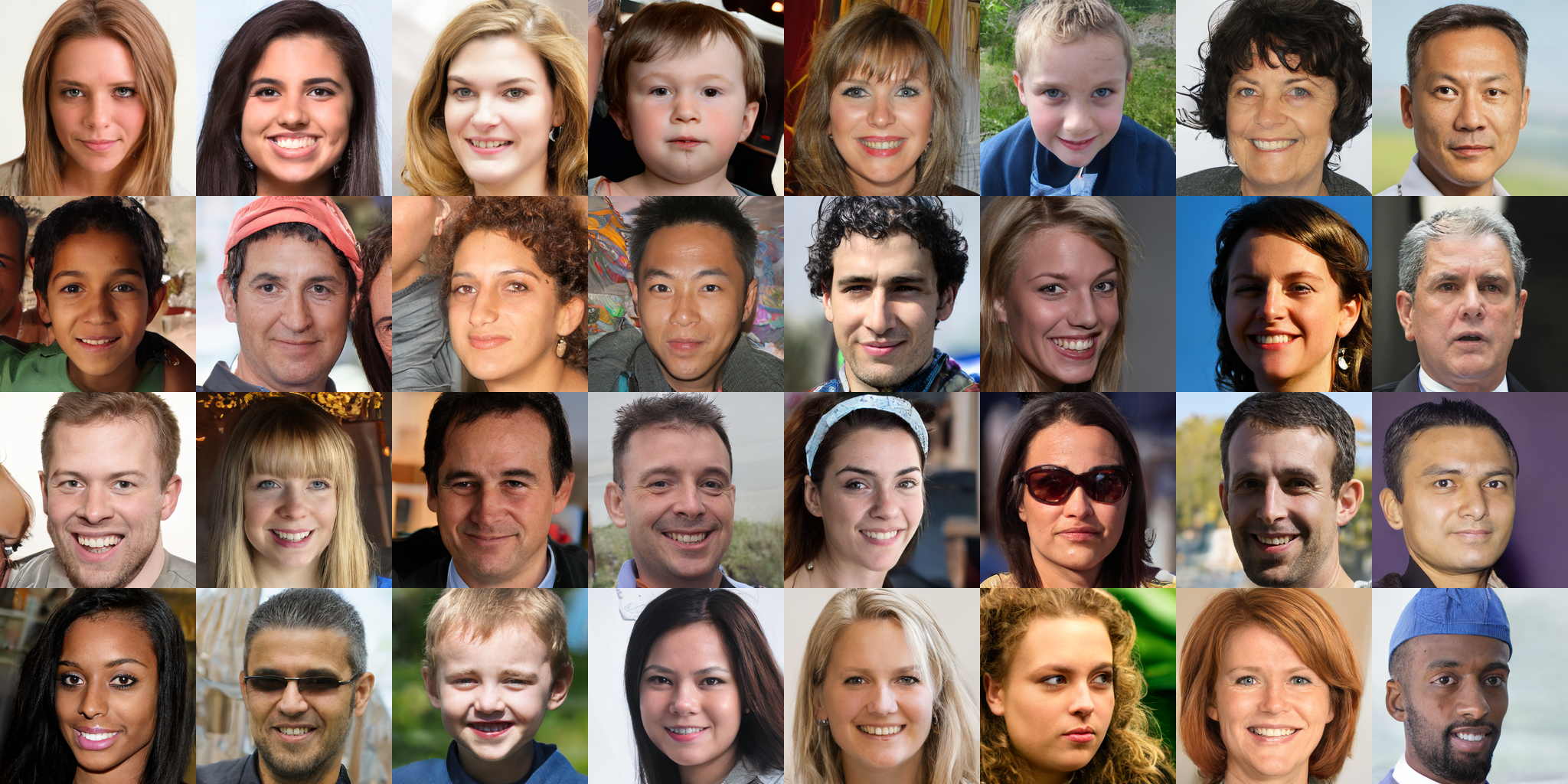}
    \caption{Additional uncurated generations from the FFHQ diffusion model}
    \label{fig:ffhq-extra}
\end{figure}

 \Cref{fig:ffhq-extra} provides additional generated samples from our latent diffusion model trained on FFHQ.

\tabSharedUNet
\tabViTAblation
\tabMergeLevel
\figScaleDependencyModConv
\tabTrainingResVAE

\begin{table}[t!]
    \centering
    \caption{Details of the feature-extraction module for different {\method} models.}
    \label{tab:arch-detail-feature-extraction}
    \begin{tabular}{lcccc}
        \toprule
          & \multicolumn{4}{c}{Feature extraction} \\
          \cmidrule(lr){2-5}
          Model & Input dim & Output dim & Channels & Channels multiple \\
        \midrule
        \method-S & 12 & 12 & 16 & (1, 2, 2) \\
        \method-B & 12 & 12 & 32 & (1, 2, 3) \\
        \method-M & 12 & 12 & 64 & (1, 2, 4) \\
        \method-L & 12 & 12 & 64 & (1, 2, 4) \\
        \bottomrule
    \end{tabular}
\end{table}

\begin{table}[t!]
    \centering
    \caption{Details of the feature-aggregation module for different {\method} models.}
    \label{tab:arch-detail-feature-aggregation}
    \begin{tabular}{lcccc}
        \toprule
          & \multicolumn{4}{c}{Feature aggregation} \\
          \cmidrule(lr){2-5}
          Model & Input dim & Output dim & Channels & Channels multiple \\
        \midrule
        \method-S & 36 & Latent dim ($n_z$) & 16 & (1, 2, 2)  \\
        \method-B & 36 & Latent dim ($n_z$) & 32 & (1, 2, 3) \\
        \method-M & 36 & Latent dim ($n_z$) & 32 & (1, 2, 3) \\
        \method-L & 36 & Latent dim ($n_z$) & 64 & (1, 2, 4) \\
        \bottomrule
    \end{tabular}
\end{table}

\section{Implementation details}\label{sec:imp-detail}
All models were trained with a batch size of 16 on two GPUs until the autoencoder could produce high-quality reconstructions. The training duration was 200k steps for the ImageNet 128$\times$128 models, and 100k for the ImageNet 256$\times$256 and FFHQ models. We use Adam optimizer \citep{Adam} with a learning rate of $10^{-4}$ and $(\beta_1, \beta_2) = (0.5, 0.9)$. The details of the model architecture for feature-extraction and feature-aggregation modules are given in \Cref{tab:arch-detail-feature-extraction,tab:arch-detail-feature-aggregation}. Our implementation of the UNet used for feature extraction and aggregation closely follows the ADM model \citep{dhariwalDiffusionModelsBeat2021} without spatial down/upsampling layers. The decoder in {\method} exactly follows the implementation of the decoder from Stable Diffusion VAE \cite{rombachHighResolutionImageSynthesis2022}, except for the SMC experiment. For training the latent diffusion and the standard VAE models, we closely follow \citet{rombachHighResolutionImageSynthesis2022} to ensure a fair comparison.


\section{Pseudocode for different LiteVAE blocks}\label{sec:pseudocode} In this section, we present additional pseudocode for various {\method} components. The core element of {\method} is the Haar wavelet transform, which can be implemented in PyTorch as shown below:

\begin{mintedbox}{python}[]
class HaarTransform(nn.Module):
  def __init__(self, level=3, mode="symmetric", with_grad=False) -> None:
    super().__init__()
    self.wavelet = pywt.Wavelet("haar")
    self.level = level
    self.mode = mode
    self.with_grad = with_grad

  def dwt(self, x, level=None):
    with torch.set_grad_enabled(self.with_grad):
      level = level or self.level
      x_low, *x_high = ptwt.wavedec2(
        x.float(), 
        wavelet=self.wavelet, 
        level=level, 
        mode=self.mode,
      )
      x_combined = torch.cat(
        [x_low, x_high[0][0], x_high[0][1], x_high[0][2]], dim=1
      )
      return x_combined

  def idwt(self, x):
    with torch.set_grad_enabled(self.with_grad):
      x_low, x_high = x[:, :3], x[:, 3:]
      x_high = torch.chunk(x_high, 3, dim=1)
      x_recon = ptwt.waverec2([x_low.float(), x_high.float()], wavelet=self.wavelet)
      return x_recon

  def forward(self, x, inverse=False):
    if inverse:
      return self.idwt(x)
    return self.dwt(x)
  \end{mintedbox}

The PyTorch implementation of the self-modulated convolution block introduced in \Cref{sec:smc} is provided below:

\begin{mintedbox}{python}[]
class SMC(nn.Module):
  def __init__(
    self,
    in_channels: int,
    out_channels: int = None,
    kernel_size: int = 3,
    stride: int = None,
    padding: int = None,
    bias: bool = True,
  ):
    super().__init__()

    # setting the default values
    out_channels = out_channels or in_channels
    padding_ = int(kernel_size // 2) if padding is None else padding
    stride_ = 1 if stride is None else stride

    self.padding = padding_

    self.conv = nn.Conv2d(
      in_channels,
      out_channels,
      kernel_size=kernel_size,
      padding=padding_,
      stride=stride_,
      bias=bias,
    )
  
    self.gain = nn.Parameter(torch.ones(1))
    self.scales = nn.Parameter(torch.ones(in_channels))
  
  def forward(self, x: torch.Tensor) -> torch.Tensor:
    scales = self.scales.expand(x.shape[0], -1)
    out = modulated_conv2d(
      x=x,
      w=self.conv.weight,
      s=scales,
      padding=self.padding,
      input_gain=self.gain,
    )
    if self.conv.bias is not None:
      out = out + self.conv.bias.view(1, -1, 1, 1)
    return out
\end{mintedbox}

Next, we present the code for the residual blocks utilized in the {\method} UNet networks:

\begin{mintedbox}{python}[]
class ResBlock(nn.Module):
  def __init__(
    self,
    in_channels: int,
    dropout: float = 0.0,
    out_channels: int = None,
    use_conv: bool = False,
    activation: str = "swish",
    norm_num_groups: int = 32,
    scale_factor: float = 1,
  ):

  super().__init__()
  self.in_channels = in_channels
  self.out_channels = out_channels or in_channels

  self.norm_in = GroupNorm(in_channels, norm_num_groups)
  self.act_in = SiLU()
  self.conv_in = ConvLayer2D(in_channels, out_channels, 3)
  self.norm_out = GroupNorm(out_channels, norm_num_groups)
  self.act_out = SiLU()
  self.dropout = Dropout(dropout)
  self.conv_out = ConvLayer2D(out_channels, 3)

  if self.out_channels == in_channels:
    self.skip_connection = Identity()
  elif use_conv:
    self.skip_connection = ConvLayer2D(in_channels, out_channels, 3)
  else:
    self.skip_connection = ConvLayer2D(in_channels, out_channels, 1)
  self.scale_factor = scale_factor

def forward(self, x):
  # input layers
  h = self.norm_in(x)
  h = self.act_in(h)
  h = self.conv_in(h)
  # output layers
  h = self.norm_out(h)
  h = self.act_out(h)
  h = self.dropout(h)
  h = self.conv_out(h)
  return (self.skip_connection(x) + h) / self.scale_factor


class ResBlockWithSMC(nn.Module):
  def __init__(
    self,
    in_channels: int,
    dropout: float = 0.0,
    out_channels: int = None,
    use_conv: bool = False,
    activation: str = "swish",
    norm_num_groups: int = 32,
    scale_factor: float = 1,
  ):

    super().__init__()
    self.in_channels = in_channels
    self.out_channels = out_channels or in_channels

    self.act_in = SiLU()
    self.conv_in = SMC(in_channels, out_channels, 3)
    self.act_out = SiLU()
    self.dropout = Dropout(dropout)
    self.conv_out = SMC(out_channels, 3)

    if self.out_channels == in_channels:
      self.skip_connection = Identity()
    elif use_conv:
      self.skip_connection = ConvLayer2D(in_channels, out_channels, 3)
    else:
      self.skip_connection = ConvLayer2D(in_channels, out_channels, 1)
    self.scale_factor = scale_factor

  def forward(self, x):
    # input layers
    h = self.act_in(x)
    h = self.conv_in(h)
    # output layers
    h = self.act_out(h)
    h = self.dropout(h)
    h = self.conv_out(h)
    return (self.skip_connection(x) + h) / self.scale_factor


class MidBlock2D(nn.Module):
  def __init__(
    self,
    in_channels: int,
    out_channels: int,
    dropout: float = 0.0,
    use_smc: bool = True,
  ) -> None:
    super().__init__()
    resblock_class = ResBlockWithSMC if use_smc else ResBlock
    self.res0 = resblock_class(
      in_channels=in_channels,
      out_channels=out_channels,
      dropout=dropout,
    )
    self.res1 = resblock_class(
      in_channels=out_channels,
      out_channels=out_channels,
      dropout=dropout,
    )
  def forward(self, x):
    x = self.res0(x)
    x = self.res1(x)
    return x
\end{mintedbox}

Additionally, the feature-extraction and feature-aggregation UNets can be implemented as follows:

\begin{mintedbox}{python}[]
class LiteVAEUNetBlock(nn.Module):
  def __init__(
    self, 
    in_channels: int, 
    out_channels: int, 
    model_channels: int, 
    ch_multiplies: list[int] = [1, 2, 4], 
    num_res_blocks: int = 2,
    use_smc: bool = False,
  ):
    super().__init__()
    self.in_layer = ConvLayer2D(in_channels, model_channels, 3)
    self.out_layer = ConvLayer2D(model_channels, out_channels, 3)

    resblock_class = ResBlockWithSMC if use_smc else ResBlock

    # -----------------------------------------------------------------
    # UNet encoder path
    # -----------------------------------------------------------------
    channel = model_channels
    in_channel_list = [model_channels]
    self.encoder_blocks = []
    for level, ch_mult in enumerate(ch_multiplies):
      for i in range(num_res_blocks):
        self.encoder_blocks.append(
          resblock_class(
            in_channels=channel,
            out_channels=model_channels * ch_mult
          )
        )
        channel = model_channels * ch_mult
        in_channel_list.append(channel)
    self.encoder_blocks = nn.ModuleList(self.encoder_blocks)
    # -----------------------------------------------------------------
    # UNet middle block
    # -----------------------------------------------------------------
    self.mid_block = MidBlock2D(
      in_channels=channel, 
      out_channels=channel, 
      embed_channels=0, 
      legacy=legacy
    )
    # -----------------------------------------------------------------
    # UNet decoder path
    # -----------------------------------------------------------------
    self.decoder_blocks = []
    for level, ch_mult in reversed(list(enumerate(ch_multiplies))):
      for i in range(num_res_blocks):
        self.decoder_blocks.append(
          resblock_class(
            in_channels=channel + in_channel_list.pop(),
            out_channels=model_channels * ch_mult
          )
        )
        channel = model_channels * ch_mult
    self.decoder_blocks = nn.ModuleList(self.decoder_blocks)

  def forward(self, x):
    x = self.in_layer(x)
    skip_features = [x]
    # the encoder path
    for enc_block in self.encoder_blocks:
      x = enc_block(x)
      skip_features.append(x)
    # the middle block
    x = self.mid_block(x)
    # the decoder path
    for dec_block in self.decoder_blocks:
      x_cat = torch.cat([x, skip_features.pop()], dim=1)
      x = dec_block(x_cat)
    return self.out_layer(x)
\end{mintedbox}

The {\method} encoder can be implemented as shown below:
\begin{mintedbox}{python}[]
class LiteVAEEncoder(nn.Module):
  def __init__(
    self,
    in_channels: int,
    out_channels: int,
    wavelet_fn: HaarTransform,
    feature_extractor_params: dict,
    feature_aggregator_params: dict,
  ):
    super().__init__()
    self.wavelet_fn = wavelet_fn
    self.feature_extractor_L1 = LiteVAEUNetBlock(
      in_channels, in_channels, **feature_extractor_params
    )
    self.feature_extractors_L2 = LiteVAEUNetBlock(
      in_channels, in_channels, **feature_extractor_params
    )
    self.feature_extractor_L3 = LiteVAEUNetBlock(
      in_channels, in_channels, **feature_extractor_params
    )
    out_channels = out_channels * 2 # for VAE mean and log_var
    aggregated_channels = in_channels * 3
    self.feature_aggregator = LiteVAEUNetBlock(
      aggregated_channels, out_channels, **feature_aggregator_params
    )
    self.downsample_block_L1 = Downsample2D(in_channels, scale_factor=4)
    self.downsample_block_L2 = Downsample2D(in_channels, scale_factor=2)

  def forward(self, image):
    dwt_L1 = self.wavelet_fn.dwt(image, level=1) / 2
    dwt_L2 = self.wavelet_fn.dwt(image, level=2) / 4
    dwt_L3 = self.wavelet_fn.dwt(image, level=3) / 8
    features_L1 = self.downsample_block_L1(
        self.feature_extractor_L1(dwt_L1)
      )
    features_L2 = self.downsample_block_L2(
        self.feature_extractor_L1(dwt_L2)
      )
    features_L3 = self.feature_extractor_L3(dwt_L3)
    dwt_features = [features_L1, features_L2, features_L3] 
    latent = self.feature_aggregator(torch.cat(features, dim=1))
    return latent
\end{mintedbox}

Finally, the code for {\method} is also provided below.

\begin{mintedbox}{python}[]
class LiteVAE(nn.Module):
  def __init__(
    self,
    encoder: LiteVAEEncoder,
    decoder: SDVAEDecoder,
    config: DictConfig,
    output_type: str = "image",
  ):
    super().__init__()
    assert output_type in ["image", "wavelet"]
    self.encoder = encoder
    self.decoder = decoder
    self.wavelet_fn = encoder.wavelet_fn
    self.output_type = output_type

    pre_channels = config.latent_dim * 2 # for VAE mean and log_var
    post_channels = config.latent_dim
    if config.get("use_1x1_conv", False):
      self.pre_conv = nn.Conv2d(pre_channels, pre_channels, 1)
      self.post_conv = nn.Conv2d(post_channels, post_channels, 1)
    else:
      self.pre_conv = nn.Identity()
      self.post_conv = nn.Identity()

  def encode(self, image):
    return self.pre_conv(self.encoder(image))

  def decode(self, latent):
  latent = self.post_conv(latent)
    if self.output_type == "image":
      image_recon = self.decoder(latent)
      wavelet_recon = self.wavelet_fn.dwt(image_recon, level=1) / 2
    elif self.output_type == "wavelet":
      wavelet_recon = self.decoder(latent)
      image_recon = self.wavelet_fn.idwt(wavelet_recon, level=1) * 2
    return image_recon, wavelet_recon

  def forward(self, image, sample=True):
    latent = self.encode(image)
    latent_dist = DiagonalGaussianDistribution(latent)
    latent = latent_dist.sample() if sample else latent_dist.mode()
    kl_reg = latent_dist.kl().mean()
    image_recon, wavelet_recon = self.decode(latent)
    return Dict(
      {
        "sample": image_recon,
        "wavelet": wavelet_recon,
        "latent": latent,
        "kl_reg": kl_reg,
        "latent_dist": latent_dist,
      }
    )
  \end{mintedbox}


\end{document}